\documentclass[letterpaper]{article} 
\usepackage{aaai25}  
\usepackage{times}  
\usepackage{helvet}  
\usepackage{courier}  
\usepackage[hyphens]{url}  
\usepackage{graphicx} 
\usepackage{natbib}  
\usepackage{caption} 
\usepackage{algorithm}
\usepackage{algorithmic}

\usepackage{newfloat}
\usepackage{listings}
\DeclareCaptionStyle{ruled}{labelfont=normalfont,labelsep=colon,strut=off} 
\floatstyle{ruled}
\newfloat{listing}{tb}{lst}{}
\floatname{listing}{Listing}
\nocopyright 

\usepackage{amsmath}
\usepackage{amssymb}
\usepackage{booktabs}
\usepackage{colortbl}
\usepackage{multirow}
\usepackage{pgfplots}
\usepackage{subcaption}
\usepackage{tikz}

\newcommand{\norm}[1]{\left\lVert#1\right\rVert}
\newcommand{\greyrule}{\arrayrulecolor{black!30}\midrule\arrayrulecolor{black}}
\DeclareMathOperator*{\argmax}{\arg\!\max}
\DeclareMathOperator*{\argmin}{\arg\!\min}

\title{My Publication Title --- Multiple Authors}
\title{Interpretable Multi-task Learning with Shared Variable Embeddings}
\author {
    Maciej {\.Z}elaszczyk\textsuperscript{\rm 1},
    Jacek Ma{\'n}dziuk\textsuperscript{\rm 1,\rm 2}
}
\affiliations {
    \textsuperscript{\rm 1}Faculty of Mathematics and Information Science, Warsaw University of Technology, Poland\\
    \textsuperscript{\rm 2}Faculty of Computer Science, AGH University of Krakow, Poland\\
    m.zelaszczyk@mini.pw.edu.pl, jacek.mandziuk@pw.edu.pl 
}

\usepackage{bibentry}

\begin{document}

\maketitle

\begin{abstract}
This paper proposes a general interpretable predictive system with shared information. The system is able to perform predictions in a multi-task setting where distinct tasks are not bound to have the same input/output structure. Embeddings of input and output variables in a common space are obtained, where the input embeddings are produced through attending to a set of shared embeddings, reused across tasks. All the embeddings are treated as model parameters and learned. The approach is distinct from existing vector quantization methods. Specific restrictions on the space of shared embedings and the sparsity of the attention mechanism are considered. Experiments show that the introduction of shared embeddings does not deteriorate the results obtained from a vanilla variable embeddings method. We run a number of further ablations. Inducing sparsity in the attention mechanism leads to both an increase in accuracy and a significant decrease in the number of training steps required. Shared embeddings provide a measure of interpretability in terms of both a qualitative assessment and the ability to map specific shared embeddings to pre-defined concepts that are not tailored to the considered model. There seems to be a trade-off between accuracy and interpretability. The basic shared embeddings method favors interpretability, whereas the sparse attention method promotes accuracy. The results lead to the conclusion that variable embedding methods may be extended with shared information to provide increased interpretability and accuracy.
\end{abstract}

%

\section{Introduction}
\label{sec:introduction}
The ability to extract common information from varied settings has long been one of the central challenges in machine learning. The degree to which the considered domains differ and the complexity of the domains themselves has grown considerably over time. Artificial neurons \cite{McCulloch1943} linked in a physical perceptron model \cite{Rosenblatt1958} were used to distinguish, through a weight update procedure, the side on which a punch card had been marked. CNNs \cite{Fukushima1980, LeCun1989} are able to identify similar patterns in distinct areas of an image. Word embeddings \cite{Bengio2000, Mikolov2013} have been used to obtain representations fusing information from different contexts. Reinforcement learning agents achieved relatively high performance in a number of Atari games without changes to the model architecture \cite{Mnih2015}. Generative methods, such as diffusion models \cite{Rombach2022} produce high-fidelity imagery based on information from a provided text prompt.

\textit{Multi-task learning} (MTL) is a specific area of machine learning concerned with approaches that attempt to simultaneously solve more than one task \cite{Caruana1993, Caruana1994, Caruana1996, Thrun1996, Caruana1997}. With the evolution of deep learning architectures over the years, we have seen a sharp increase in the interest in MTL, e.g. in hard parameter sharing methods \cite{Hu2021, Cui2021}, soft parameter sharing methods \cite{Misra2016, Gao2019}, decoder models \cite{Bruggemann2021, Ye2023}. MTL also has strong links to \textit{self-supervised learning} (SSL) — a learning paradigm where models can be pretrained on unlabelled data in order to improve their performance or data efficiency on downstream tasks \cite{Mikolov2013, Chen2020, Zbontar2021, Bardes2022, Assran2023}.

MTL and SSL can be interpreted as settings in which knowledge about one task facilitates the learning of another one. This is usually done via a choice of tasks that share significant structure, e.g. semantic segmentation, human parsing, monocular depth estimation, etc. This is in stark contrast to real-world prediction scenarios, where typically no a priori structure is given. The difficulty in obtaining meaningful structure from unrelated tasks has led research on MTL to mostly focus on related tasks, even though there is a body of work suggesting that solving not obviously related tasks may actually be helpful in making sound predictions \cite{Mahmud2007, Meyerson2019}.

A specific line of investigation for tabular data postulates casting variables associated with unrelated tasks into a shared embedding space, on top of simply measuring the value of these variables \cite{Meyerson2021}. This can also be understood as associating with each variable a $(\text{key}, \text{value})$ pair, where the key is a \textit{variable embedding}. This draws inspiration from word embeddings \cite{Bengio2000}, attention \cite{Bahdanau2015, Vaswani2017} and key-value retrieval methods \cite{Graves2014, Graves2016, Goyal2021}. Such an approach affords to perform classification or regression and to tackle distinct tasks with different numbers of inputs and outputs in order to extract unobvious common information. It does, however, require each variable to be assigned a unique embedding. This limits the ability to reason about the degree to which specific variables are similar to one another and about their shared components.

This paper aims to investigate the extent to which variable embeddings can reuse the same information and proposes a setting where each variable embedding can be represented as a reconfiguration of a common component base shared across tasks. This, in turn, allows us to link any common component to specific variables which rely on it most and to identify common concepts shared between variables.

\textbf{Motivation:} We aim to: (1) encourage information re-use by relaxing the assumption of one VE per variable, (2) facilitate interpretability in the VE setting, (3) verify whether restrictions on the shared information improve the accuracy and training efficiency of the VE method.

\textbf{Main contributions of this paper:}
\begin{itemize}
    \item Proposes a variable embedding architecture with a shared component base accessed via attention.
    \item Shows that the introduction of the shared base does not hurt performance, while allowing for a substantial reduction in training steps.
    \item Verifies that specific components from the shared base incorporate abstract intuitive concepts.
    \item Investigates specific restrictions on the form of the shared base and the attention mechanism.
    \item Identifies and investigates the trade-off between interpretability and accuracy in shared embedding systems.
\end{itemize}

\section{Background}
\label{sec:background}
We consider a setting with $T$ tasks $\{\left(\mathbf{x}_{t}, \mathbf{y}_{t}\right) \}_{t = 1}^{T}$, where task $t$ has $n_{t}$ input variables $\left[x_{t1}, \dots, x_{tn}\right] = \mathbf{x}_{t} \in \mathbb{R}^{n_{t}}$ and $m_{t}$ output variables $\left[y_{t1}, \dots, y_{tn}\right] = \mathbf{y}_{t} \in \mathbb{R}^{m_{t}}$. Two tasks $\left(\mathbf{x}_{t}, \mathbf{y}_{t}\right)$ and $\left(\mathbf{x}_{t^{\prime}}, \mathbf{y}_{t^{\prime}}\right)$ are said to be \textit{disjoint} if there is no overlap between their input and output variables: $\left(\{x_{ti}\}_{i = 1}^{n_{t}} \cup\{y_{tj}\}_{j = 1}^{m_{t}}\right) \cap \left(\{x_{t^{\prime}i}\}_{i = 1}^{n_{t^{\prime}}} \cup\{y_{t^{\prime}j}\}_{j = 1}^{m_{t^{\prime}}}\right) = \varnothing$.

The notion of word embeddings \cite{Bengio2000} can be extended to \textit{variable embeddings} (VEs) \cite{Meyerson2021} by treating the $i$-th variable as being associated with two elements:
\begin{itemize}
    \item A specific \textit{variable embedding} $\mathbf{z}_{i} \in \mathbb{R}^{C}$, which can be interpreted as the \textit{name} or \textit{key} of that variable. $C$ is the dimensionality of the embedding.
    \item A specific scalar \textit{value} $v_{i} \in \mathbb{R}$.
\end{itemize}
In particular, much like word embeddings, variable embeddings do not necessarily have to be specified in advance as they can be treated as parameters of a model and learned.

Let us describe a \textit{prediction task} $\left(\mathbf{x}, \mathbf{y}\right) = \left(\left[x_{1}, \dots, x_{n}\right]\right.$, $\left.\left[y_{1}, \dots, y_{m}\right]\right)$. The goal is to predict the values of \textit{target variables} $\{y_{j}\}_{j = 1}^{m}$ (output) from the values of \textit{observed variables} $\{x_{i}\}_{i = 1}^{n}$ (input). Notably, a \textit{classification task} is a special case of a prediction task with target variables restricted to one-hot encodings.

A \textit{predictor} $\Omega$ is a function which maps between observed and target variables. Let $\mathbf{z}_{i}$ and $\mathbf{z}_{j}$ be the variable embeddings of $x_{i}$ and $y_{j}$. An MTL predictor can then be defined as:
\begin{equation}
    \mathbb{E}\left[y_{j} \vert \mathbf{x}\right] = \Omega\left(\mathbf{x}, \{\mathbf{z}_{i}\}_{i = 1}^{n}, \mathbf{z}_{j}\right)
\end{equation}
$\Omega$ is shared across tasks to extract common knowledge and the tasks themselves are identified via their variable embeddings.
A particular form of $\Omega$ is obtained by expressing the predictor via function composition:
\begin{equation}
    \Omega\left(\mathbf{x}, \{\mathbf{z}_{i}\}_{i = 1}^{n}, \mathbf{z}_{j}\right) = g\left(\sum_{i = 1}^{n}{f(x_{i}, \mathbf{z}_{i})}, \mathbf{z}_{j}\right)
\end{equation}
where $f$ is an \textit{encoder}, $g$ is a \textit{decoder}, and there is an implicit assumption that the ordering of observed variables does not matter. $f: \mathbb{R}^{C + 1} \to \mathbb{R}^{M}$, $g: \mathbb{R}^{M + C} \to \mathbb{R}$, where $M$ is the dimension of the latent space to which the encoder maps. $f$ and $g$ can be approximated with neural networks $f_{\theta_{f}}$ and $g_{\theta_{g}}$ where $\theta_{f}$ and $\theta_{g}$ are parameters learned by gradient descent.

The decoder can be further decomposed for computational efficiency:
\begin{equation}
\label{eq:second-decomposition}
    \mathbb{E}\left[y_{j} \vert \mathbf{x}\right] = g_{2}\left(g_{1}\left(\sum_{i = 1}^{n}{f(x_{i}, \mathbf{z}_{i})}\right), \mathbf{z}_{j}\right)
\end{equation}
where $g_{1}$ is the initial decoder which is independent of the target variable being predicted, while $g_{2}$ is the final decoder conditioned on the target variable's embedding. This allows $g_{1}$ to learn transformations of the observed variables not dependent on the specific output variable. Also, $g_{1}\left(\sum_{i = 1}^{n}{f(x_{i}, \mathbf{z}_{i})}\right)$ can be pre-computed ahead of specific predictions for a given target variable.

As far as specific choices of architectures of the encoder and decoders are concerned, we follow the setup presented in the Traveling Observer Model (TOM) \cite{Meyerson2021} where the conditioning on variable embeddings is done via FiLM layers \cite{Perez2018}. The general motivation behind VEs is discussed in Appendix \ref{app:variable-embeddings}.

The described procedure shows specific advantages, e.g. the possibility to handle tasks with different dimensions of input and output spaces, the ability to recover structure on small-scale problems, and relatively good performance on a range of tasks. On the flip side, it does not reuse the obtained embeddings between variables and it does not lend itself readily to interpretation for real-world classification datasets. In principle, a variable embedding is obtained for each observed and target variable, so the embeddings can be compared in their common space or projected into a lower-dimension space for visualization using methods such as t-SNE \cite{Hinton2002} or UMAP \cite{McInnes2018}. In reality, however, this turns out to be problematic for more complex data. For instance, for the real world dataset of UCI-121 \cite{Delgado2014, Kelly2023}, the vanilla variable embeddings approach produces embeddings which seem to differentiate between the observed variables, common target variables and uncommon target variables \cite{Meyerson2021}, but we do not have any more information on the relations between the variables themselves.

\section{Method}
\label{sec:method}
\begin{figure*}[ht!]
\vskip 0.2in
\begin{center}
\centerline{\includegraphics[width=1.6\columnwidth]{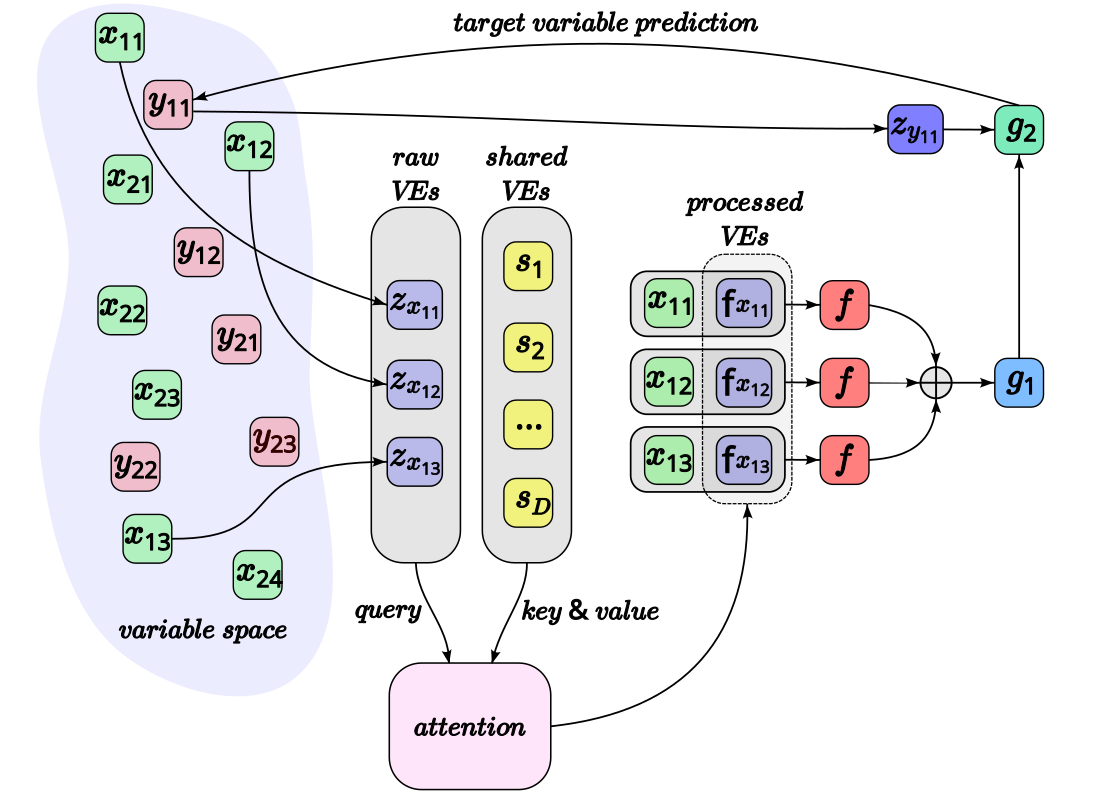}}
\caption{The overview of the \textit{shared variable embeddings} method. The \textit{variable space} contains both the \textit{observed} and \textit{target} variables which are associated with their learnable \textit{variable embeddings} (VEs). The observable variables are first linked to \textit{raw VEs} which are used as \textit{queries} in the attention mechanism. A separate set of \textit{shared VEs} plays the role of both \textit{keys} and \textit{values}. The \textit{processed VEs} are the output of attention. Together with the corresponding variable values they are processed, each (value, VE) pair separately, by the \textit{encoder}. The outputs of the encoder are summed and passed to the initial \textit{decoder}. The target variable of interest is directly linked to its VE and this VE is passed with the output of the initial decoder to the final decoder to actually perform the prediction of the value of the target variable of interest. Additional details of the architecture are available in Appendix \ref{app:architecture} (Figure~\ref{fig:architecture}). The differences between our method and VQ-VAE \cite{vandenOord2017} are highlighted in Appendix \ref{app:vq-vae}.}
\label{fig:method}
\end{center}
\vskip -0.2in
\end{figure*}

In order to encourage the reuse of information between the variables and to increase the interpretability of the approach, we propose \textit{shared variable embeddings}, selectively used for each observed variable. The outline of our method is shown in Figure \ref{fig:method}.

\subsection{Shared variable embeddings}
Let us consider $N$ observed variables and a set of $D$ shared embeddings $\{\mathbf{s}_{k}\}_{k = 1}^{D}$, with $\mathbf{s}_{k} \in \mathbb{R}^{C}$. The associated \textit{shared embedding matrix} is $\mathbf{S}_{D \times C}$, where $C$ is the dimension of both the embedding space of observed variables and the shared embedding space. In order to enforce the reuse of information between variable embeddings, we would like $D \ll N$. We relate the raw (initial) variable embeddings to the shared embeddings via attention. The increase in model parameters is motivated in Appendix \ref{app:parameters-increase}.

For the \textit{raw variable embedding matrix} $\mathbf{Z}_{N \times C}$, we follow the standard attention procedure \cite{Vaswani2017}:
\begin{equation}
    A(\mathbf{Q}, \mathbf{K}, \mathbf{V}) = softmax\left( \frac{\mathbf{Q}\mathbf{K}^{T}}{\sqrt{d_{k}}} \right)\mathbf{V} 
\end{equation}
where $\mathbf{Q}$, $\mathbf{K}$, $\mathbf{V}$ can be interpreted as the matrices of \textit{queries}, \textit{keys} and \textit{values}, respectively, and $d$ is the dimensionality of both the queries and keys. In our specific case, we will apply cross-attention and the initial variable embeddings can be assigned the role of the queries, while the shared variable embeddings are assigned the roles of both the keys and the values:
\begin{equation}
    A(\mathbf{Z}, \mathbf{S}, \mathbf{S}) = softmax\left( \frac{\mathbf{Z}\mathbf{S}^{T}}{\sqrt{C}} \right)\mathbf{S} 
\end{equation}
The output of this procedure is the \textit{processed variable embedding matrix} $\mathbf{F}_{N \times C}$, where each \textit{processed variable embedding} $\mathbf{f}_{i}$ is a linear combination of all the shared embeddings, weighted by their similarity score to the \textit{raw variable embedding} $\mathbf{z}_{i}$. Similarly to standard variable embeddings, shared variable embeddings can be either handcrafted or learned as model parameters.

Once the processed embedding has been obtained it can be substituted into Eq. \ref{eq:second-decomposition} to get:
\begin{equation}
\label{eq:shared}
    \mathbb{E}\left[y_{j} \vert \mathbf{x}\right] = g_{2}\left(g_{1}\left(\sum_{i = 1}^{n}{f(x_{i}, \mathbf{f}_{i})}\right), \mathbf{z}_{j}\right)
\end{equation}
An important distinction between standard and shared variable embeddings is that the standard ones are inextricably tied to a specific variable from a specific dataset, while in our shared version each shared embedding is not directly linked to one specific variable from a given dataset and can be potentially reused between variables and datasets.

\subsection{Training}
The proposed model is trained end-to-end with stochastic gradient descent. For one training step, a two-fold procedure follows. First, a task is sampled from the distribution of overall tasks considered. Second, using the dataset associated with the sampled task, a sample of training examples is drawn. For each of these examples, standard variable embeddings are obtained for each of the observed and target variables. Those for the observed variables are passed through the attention mechanism to use the shared embeddings and obtain the processed variable embeddings. Such embeddings are then passed through the encoder/decoder architecture in order to obtain predictions for each target variable. These predictions are used to calculate the squared hinge loss:
\begin{equation}
    L(\hat{\mathbf{y}}, \mathbf{t}) = \sum_{j = 1}^{m}{\max\left(0, 1 - t_{j} \cdot \hat{y_{j}}\right)^{2}} 
\end{equation}
where $t_{j}$ is a $+1/-1$ encoding of the actual target and $\hat{y}_{j}$ is the prediction of the value of the $j$-th target variable for the given task obtained from the encoder/decoder architecture with shared variable embeddings as in Eq. \ref{eq:shared}. Details of the hinge loss are discussed in Appendix \ref{app:hinge-loss}.

\subsection{Imposing independence of shared variable embeddings through additional structure}
One question that can be asked of the shared embedding matrix $\mathbf{S}_{D \times C}$ is that of structure. In particular, it could be argued that the learning process does not explicitly require the shared embeddings $\{\mathbf{s}_{k}\}_{k = 1}^{D}$ to be \textit{independent} from one another. We consider different notions of independence and several approaches to encourage it in the shared embeddings:
\begin{itemize}
    \item \textbf{Orthogonalization:} encouraging $\mathbf{S}$ to consist of \textit{orthonormal} vectors.
    \item \textbf{Stable rank:} nudging $\mathbf{S}$ to have high rank.
    \item \textbf{Von Neumann entropy:} optimizing for vectors in $\mathbf{S}$ to be independent from the point of view of information theory.
    \item \textbf{Sparse attention:} adding sparsity to the attention mechanism.
\end{itemize}
The specific details of all these approaches are discussed in Appendix \ref{app:restrictions}.

\begin{table}
\caption{Best classification accuracy for variable embedding methods on the UCI-121 test set.}
\label{tab:accuracy-overall}
\vskip 0.15in
\begin{center}
\begin{small}
\begin{sc}
\begin{tabular}{lcc}
\toprule
Method & Accuracy & No fine-tuning? \\
\midrule
vanilla & 81.5 & $\times$ \\
shared embedding & 81.5 & $\surd$ \\
\textbf{1.05-entmax} & \textbf{81.9} & $\surd$ \\
stable rank, $\alpha_{\text{sr}} =$ 0.05 & 80.6 & $\surd$ \\
\bottomrule
\end{tabular}
\end{sc}
\end{small}
\end{center}
\vskip -0.1in
\end{table}

\section{Experiments}
\label{sec:experiments}
We validate the ability of shared variable embeddings to solve real-life classification tasks and to help in interpretability on the UCI-121 dataset \cite{Delgado2014, Kelly2023} (Appendix \ref{app:uci-121}). In the experiments, we use the hyperparameter values and the learning setup from \cite{Meyerson2021}. Information on code and data availability is included in Appendix \ref{app:code}. For the shared variable embeddings, we choose $D = C = 128$. We provide quantitative comparisons of the proposed method against a strong baseline: the variable embedding method without shared embeddings. We also consider the results with restrictions on the shared embedding matrix and on the attention mechanism, as proposed in Section~\ref{sec:method}. We provide qualitative comparisons for the basic version of our method and for its configurations with constraints on the shared embedding matrix and with sparse attention. Additionally, we report the results of extensive ablations (Appendix \ref{app:hyperparameter}), per-task metrics (Appendix \ref{app:tasks}), experiments on additional datasets (Appendix \ref{app:additional-datasets}) and training time (Appendix \ref{app:training-time}).

\subsection{Classification capability}
\label{sec:classification-capability}
Results in terms of best test set accuracy are presented in Table \ref{tab:accuracy-overall}. We use the vanilla variable embedding method without shared embeddings \cite{Meyerson2021} as a strong baseline. Quantitative assessment shows that the shared embedding approach is able to achieve a classification accuracy in the range of the results from the baseline. At the same time, the $1.05$-entmax sparse attention method is able to surpass the accuracy levels of both the baseline and the shared embedding method with full attention. Notably, the baseline requires additional fine-tuning on each of the 121 datasets while our approaches do not.
\begin{table}
\centering
\caption{Classification accuracy (ACC) for variable embedding methods with orthogonalization (left), stable rank (middle), von Neumann entropy (right) on the UCI-121 test set.}
\label{tab:accuracy-orth-sr-vn}
\begin{center}
\begin{small}
\begin{sc}
\begin{tabular}{lc|lc|lc}
\toprule
$\alpha_{\text{orth}}$ & ACC & $\alpha_{\text{sr}}$ & ACC & $\alpha_{\text{vN}}$ & ACC\\
\midrule
0 & \textbf{75.5} & 0.01 & 79.9 & 0.001 & 68.6\\
0.1 & 74.5 & 0.04 & 80.3 & 0.01 & 67.8\\
1 & 74.3 & 0.05 & 80.3 & 0.05 & \textbf{71.9}\\
10 & 72.4 & 0.06 & \textbf{80.7} & 0.5 & 69.6 \\
100 & 74.3 & 0.1 & 79.0 & & \\
1000 & 71.4 & 0.5 & 79.7 & & \\
& & 1 & 76.2 & & \\
\bottomrule
\end{tabular}
\end{sc}
\end{small}
\end{center}
\end{table}

\begin{figure}
\vskip 0.2in
\begin{center}
\centerline{
\resizebox{0.35\textwidth}{0.25\textwidth}{
    \begin{tikzpicture}
    \begin{axis}[
        xlabel={training step (k)},
        ylabel={test accuracy},
        xmin=0, xmax=18,
        ymin=0.3, ymax=1,
        xtick={0,1,...,18},
        ytick={0.3,0.4,...,1.0},
        x tick label style={
            rotate=-90,
        },
        y tick label style={
            /pgf/number format/.cd,
            fixed,
            fixed zerofill,
            precision=1,
            /tikz/.cd
        },
        legend pos=south east,
        ymajorgrids=true,
        grid style=dashed,
        every axis plot/.append style={ultra thick},
    ]
    
    \addplot[
        color=teal,
        ]
        coordinates {
        (0.0, 0.343696931360369)
        (0.1, 0.493561728041365)
        (0.2, 0.506179301337351)
        (0.3, 0.5337230200436)
        (0.4, 0.554640776363893)
        (0.5, 0.584171645603216)
        (0.6, 0.597988859697268)
        (0.7, 0.626152177728543)
        (0.8, 0.644497231403579)
        (0.9, 0.660859540746425)
        (1.0, 0.672752688599386)
        (1.1, 0.67747312346612)
        (1.2, 0.699466904020406)
        (1.3, 0.707846827721599)
        (1.4, 0.722279913860316)
        (1.5, 0.727974832768944)
        (1.6, 0.741113660948427)
        (1.7, 0.740120813325697)
        (1.8, 0.744855944358841)
        (1.9, 0.743942857534664)
        (2.0, 0.746963335791067)
        (2.1, 0.745697244355482)
        (2.2, 0.748295027600739)
        (2.3, 0.752462496242645)
        (2.4, 0.757430362858257)
        (2.5, 0.758069639090397)
        (2.6, 0.76393361869415)
        (2.7, 0.764602230742809)
        (2.8, 0.763918895420096)
        (2.9, 0.770580663348492)
        (3.0, 0.767388695486824)
        (3.1, 0.768802484376299)
        (3.2, 0.774411537482841)
        (3.3, 0.775225024459611)
        (3.4, 0.778007328460978)
        (3.5, 0.780948034323181)
        (3.6, 0.781097400333486)
        (3.7, 0.773755374972235)
        (3.8, 0.775090742747944)
        (3.9, 0.779090500088312)
        (4.0, 0.780133978286496)
        (4.1, 0.781462698464013)
        (4.2, 0.781350427765567)
        (4.3, 0.782649967601251)
        (4.4, 0.781590548946483)
        (4.5, 0.792274810077893)
        (4.6, 0.785975267215807)
        (4.7, 0.786984527669569)
        (4.8, 0.786591192068172)
        (4.9, 0.79470654101886)
        (5.0, 0.795556783462379)
        (5.1, 0.784981624710032)
        (5.2, 0.795001598365231)
        (5.3, 0.794190691751207)
        (5.4, 0.795865997854126)
        (5.5, 0.796844398237341)
        (5.6, 0.798280766665026)
        (5.7, 0.797666898001036)
        (5.8, 0.79835177961369)
        (5.9, 0.800278607532656)
        (6.0, 0.801881193223162)
        (6.1, 0.801153592460428)
        (6.2, 0.804488109687827)
        (6.3, 0.803027693600013)
        (6.4, 0.803287785439122)
        (6.5, 0.804952465024628)
        (6.6, 0.806268619018554)
        (6.7, 0.805203130367742)
        (6.8, 0.804833239642435)
        (6.9, 0.806109651572488)
        (7.0, 0.806327546315507)
        (7.1, 0.805317890153451)
        (7.2, 0.807572643574789)
        (7.3, 0.808982987481987)
        (7.4, 0.809768848515093)
        (7.5, 0.806343671785388)
        (7.6, 0.807424173831096)
        (7.7, 0.806363581152499)
        (7.8, 0.807402690996845)
        (7.9, 0.798485896129637)
        (8.0, 0.800712469988722)
        (8.1, 0.800400154950752)
        (8.2, 0.803093441177105)
        (8.3, 0.80339127303313)
        (8.4, 0.802485707469108)
        (8.5, 0.802079021432088)
        (8.6, 0.803167464784636)
        (8.7, 0.804127429277744)
        (8.8, 0.803008156045202)
        (8.9, 0.805013062105286)
        (9.0, 0.805921244456886)
        (9.1, 0.805673424365895)
        (9.2, 0.805993079019141)
        (9.3, 0.804725492763649)
        (9.4, 0.805064453083948)
        (9.5, 0.806810273510248)
        (9.6, 0.807768684304893)
        (9.7, 0.808000055606883)
        (9.8, 0.806949625457035)
        (9.9, 0.808472517306538)
        (10.0, 0.809486668549529)
        (10.1, 0.810543957916661)
        (10.2, 0.811658164196991)
        (10.3, 0.811212820852826)
        (10.4, 0.811180933807996)
        (10.5, 0.810863057389688)
        (10.6, 0.811024017999161)
        (10.7, 0.81316928531697)
        (10.8, 0.813195575452223)
        (10.9, 0.812203143068605)
        (11.0, 0.812127751811761)
        (11.1, 0.812324921401904)
        (11.2, 0.811461055674147)
        (11.3, 0.811362922957363)
        (11.4, 0.812020734147728)
        (11.5, 0.812316722595852)
        (11.6, 0.811630639548725)
        (11.7, 0.811720626536855)
        (11.8, 0.811787954762982)
        (11.9, 0.811753186073998)
        (12.0, 0.811690576507256)
        (12.1, 0.812249085024951)
        (12.2, 0.812065813774007)
        (12.3, 0.812454854441607)
        (12.4, 0.812720800908623)
        (12.5, 0.813444138345307)
        (12.6, 0.813781869013616)
        (12.7, 0.813815477153471)
        (12.8, 0.813651521484562)
        (12.9, 0.81354760733861)
        (13.0, 0.813766166137839)
        (13.1, 0.813834290303181)
        (13.2, 0.81371523830427)
        (13.3, 0.813918790783053)
        (13.4, 0.814074705383451)
        (13.5, 0.813787873845855)
        (13.6, 0.813905370046564)
        (13.7, 0.813265243691663)
        (13.8, 0.812850560009338)
        (13.9, 0.812912051548103)
        (14.0, 0.812894876088391)
        (14.1, 0.813039384351444)
        (14.2, 0.812921320597016)
        (14.3, 0.813041707127482)
        (14.4, 0.812903571789031)
        (14.5, 0.811948846284613)
        (14.6, 0.812036251624905)
        (14.7, 0.812128743633616)
        (14.8, 0.812950171178912)
        (14.9, 0.813091234366021)
        (15.0, 0.812957794366709)
        (15.1, 0.813078825491741)
        (15.2, 0.81300558478718)
        (15.3, 0.813768027348393)
        (15.4, 0.813273528208727)
        (15.5, 0.813081681178137)
        (15.6, 0.814120012174215)
        (15.7, 0.814120608384217)
        (15.8, 0.81357367437963)
        (15.9, 0.814642509126151)
        (16.0, 0.813565174585271)
        (16.1, 0.814712272857424)
        (16.2, 0.814745237698928)
        (16.3, 0.813829781964758)
        (16.4, 0.814110183381523)
        (16.5, 0.812783463329506)
        (16.6, 0.813263213245295)
        (16.7, 0.813809622121829)
        (16.8, 0.812791173169496)
        (16.9, 0.81301239363865)
        (17.0, 0.812928914216328)
        (17.1, 0.812928914216328)
        (17.2, 0.812928914216328)
        (17.3, 0.812928914216328)
        (17.4, 0.812928914216328)
        };
        
    \addplot[
        color=orange,
        ]
        coordinates {
        (0.0, 0.358367334391993)
        (0.1, 0.485048903257448)
        (0.2, 0.511691880586655)
        (0.3, 0.550214206742336)
        (0.4, 0.569786749789689)
        (0.5, 0.578261392545037)
        (0.6, 0.60495222753927)
        (0.7, 0.631919800454338)
        (0.8, 0.639297109817203)
        (0.9, 0.645326247636728)
        (1.0, 0.648783672768354)
        (1.1, 0.649515195799864)
        (1.2, 0.651045242101238)
        (1.3, 0.653412269182815)
        (1.4, 0.657261769673331)
        (1.5, 0.658867885216502)
        (1.6, 0.662987832872861)
        (1.7, 0.671728071985906)
        (1.8, 0.677657342541372)
        (1.9, 0.684951975395728)
        (2.0, 0.69400182362077)
        (2.1, 0.703935012577457)
        (2.2, 0.709532739806259)
        (2.3, 0.710817841590083)
        (2.4, 0.721219342671873)
        (2.5, 0.723826638805418)
        (2.6, 0.727540105083178)
        (2.7, 0.728158907160538)
        (2.8, 0.73390803679198)
        (2.9, 0.737789653585656)
        (3.0, 0.742909270290495)
        (3.1, 0.745966952641502)
        (3.2, 0.746990336160574)
        (3.3, 0.750048621965635)
        (3.4, 0.752003885895657)
        (3.5, 0.752529699482782)
        (3.6, 0.754768746633237)
        (3.7, 0.755568194772727)
        (3.8, 0.755803591484208)
        (3.9, 0.758628454919782)
        (4.0, 0.760547033881425)
        (4.1, 0.760707445590888)
        (4.2, 0.762212208407829)
        (4.3, 0.7628929274227)
        (4.4, 0.766570247528918)
        (4.5, 0.768609539485243)
        (4.6, 0.770664224265101)
        (4.7, 0.762607823810792)
        (4.8, 0.763716299198784)
        (4.9, 0.766797836775326)
        (5.0, 0.768473719620746)
        (5.1, 0.769010792193458)
        (5.2, 0.771099210559219)
        (5.3, 0.769975815675446)
        (5.4, 0.771389292434971)
        (5.5, 0.771152943412424)
        (5.6, 0.773373695803347)
        (5.7, 0.774189635665855)
        (5.8, 0.772735687668247)
        (5.9, 0.77499548046141)
        (6.0, 0.775548273606333)
        (6.1, 0.777727294745777)
        (6.2, 0.778560916047869)
        (6.3, 0.77944792984584)
        (6.4, 0.780912708230517)
        (6.5, 0.780937077193436)
        (6.6, 0.780611453046947)
        (6.7, 0.781390081573876)
        (6.8, 0.783294555354228)
        (6.9, 0.783803609379912)
        (7.0, 0.784908583326263)
        (7.1, 0.785243235952019)
        (7.2, 0.787156047101703)
        (7.3, 0.787254212554753)
        (7.4, 0.78943982673988)
        (7.5, 0.788821571540691)
        (7.6, 0.789828327870382)
        (7.7, 0.78926029487387)
        (7.8, 0.794296517654952)
        (7.9, 0.794018608071637)
        (8.0, 0.795198506685892)
        (8.1, 0.794540408171161)
        (8.2, 0.797244079021724)
        (8.3, 0.797209999783378)
        (8.4, 0.798521162487848)
        (8.5, 0.797774734686122)
        (8.6, 0.797500290160771)
        (8.7, 0.799587654092055)
        (8.8, 0.798146935985805)
        (8.9, 0.799437907700716)
        (9.0, 0.800082023192003)
        (9.1, 0.802757488132118)
        (9.2, 0.801306665174228)
        (9.3, 0.802227828824191)
        (9.4, 0.803713521081062)
        (9.5, 0.803069424644942)
        (9.6, 0.803553116164381)
        (9.7, 0.804541668700801)
        (9.8, 0.803473895460423)
        (9.9, 0.804512097944197)
        (10.0, 0.804203653237899)
        (10.1, 0.803514554104483)
        (10.2, 0.803991416066666)
        (10.3, 0.80399930682481)
        (10.4, 0.805190364060419)
        (10.5, 0.805580774669531)
        (10.6, 0.805733920917101)
        (10.7, 0.803594667167561)
        (10.8, 0.803164313798692)
        (10.9, 0.804024374727305)
        (11.0, 0.804309097566874)
        (11.1, 0.802991360211475)
        (11.2, 0.804060411087406)
        (11.3, 0.802895873146008)
        (11.4, 0.802895873146008)
        (11.5, 0.803451372296549)
        (11.6, 0.803279195988009)
        (11.7, 0.801776566386206)
        (11.8, 0.801896341209538)
        (11.9, 0.802759337940303)
        (12.0, 0.802759337940303)
        (12.1, 0.802759337940303)
        (12.2, 0.803740398425496)
        (12.3, 0.803559613301529)
        (12.4, 0.803301348838719)
        (12.5, 0.803249695946157)
        (12.6, 0.805065664912711)
        (12.7, 0.805212095735617)
        (12.8, 0.804274129485386)
        (12.9, 0.804319770733987)
        (13.0, 0.804035575732636)
        (13.1, 0.803424194341474)
        (13.2, 0.803363963757185)
        (13.3, 0.803363963757185)
        (13.4, 0.803378392743543)
        (13.5, 0.803389871164113)
        };

    \addplot[
        color=magenta,
        ]
        coordinates {
        (0.0, 0.336190089830103)
        (0.1, 0.494983286157198)
        (0.2, 0.536105741126372)
        (0.3, 0.554641945195209)
        (0.4, 0.564947687353152)
        (0.5, 0.590346706952414)
        (0.6, 0.605838428869439)
        (0.7, 0.627002964510325)
        (0.8, 0.647606103428236)
        (0.9, 0.673057122694547)
        (1.0, 0.682965925772863)
        (1.1, 0.692520761962552)
        (1.2, 0.698189344367997)
        (1.3, 0.703191094562138)
        (1.4, 0.708956896139597)
        (1.5, 0.714709344760594)
        (1.6, 0.722547796144078)
        (1.7, 0.727592311817301)
        (1.8, 0.737521408029474)
        (1.9, 0.743389646251322)
        (2.0, 0.744810697946565)
        (2.1, 0.748618208423851)
        (2.2, 0.750898972347532)
        (2.3, 0.756861033287095)
        (2.4, 0.759586564030756)
        (2.5, 0.766625859611272)
        (2.6, 0.766412809449243)
        (2.7, 0.769180160048314)
        (2.8, 0.773421472082039)
        (2.9, 0.774510435924644)
        (3.0, 0.78204968154968)
        (3.1, 0.784796074360104)
        (3.2, 0.783612108332524)
        (3.3, 0.788664790477271)
        (3.4, 0.786692441075804)
        (3.5, 0.78604747333667)
        (3.6, 0.791279989127831)
        (3.7, 0.790797203773909)
        (3.8, 0.792794379850875)
        (3.9, 0.794387296632319)
        (4.0, 0.793926846563298)
        (4.1, 0.794796611053154)
        (4.2, 0.795399924886535)
        (4.3, 0.797091271933288)
        (4.4, 0.798603790179294)
        (4.5, 0.79746512191588)
        (4.6, 0.7986967540223)
        (4.7, 0.8011652921523)
        (4.8, 0.801545492477215)
        (4.9, 0.801772478625972)
        (5.0, 0.798825186174653)
        (5.1, 0.800625103656196)
        (5.2, 0.803284650697801)
        (5.3, 0.804812399526269)
        (5.4, 0.802809485422986)
        (5.5, 0.800766018478194)
        (5.6, 0.803431341840257)
        (5.7, 0.803281491309331)
        (5.8, 0.806040391995768)
        (5.9, 0.807552105869713)
        (6.0, 0.807400763232506)
        (6.1, 0.80933920620032)
        (6.2, 0.809677020108426)
        (6.3, 0.810066355866846)
        (6.4, 0.809678405601939)
        (6.5, 0.811499640044286)
        (6.6, 0.812729897825838)
        (6.7, 0.81334055577879)
        (6.8, 0.814098205082904)
        (6.9, 0.812925947685949)
        (7.0, 0.815345734373615)
        (7.1, 0.815928371344359)
        (7.2, 0.816168815269555)
        (7.3, 0.813929080575837)
        (7.4, 0.814484123274383)
        (7.5, 0.814488389910881)
        (7.6, 0.815859635502742)
        (7.7, 0.815141063736787)
        (7.8, 0.814827618547707)
        (7.9, 0.814964948365881)
        (8.0, 0.814987670412127)
        (8.1, 0.815463149241834)
        (8.2, 0.814646591816356)
        (8.3, 0.81797903236825)
        (8.4, 0.817943255472969)
        (8.5, 0.818917907949555)
        (8.6, 0.81822946787469)
        (8.7, 0.817687316382176)
        (8.8, 0.817512091227587)
        (8.9, 0.817106052376451)
        (9.0, 0.818267571658079)
        (9.1, 0.818931201947604)
        (9.2, 0.818154041782441)
        (9.3, 0.818161028053918)
        (9.4, 0.807780603623907)
        (9.5, 0.816702083421486)
        (9.6, 0.816600603386904)
        (9.7, 0.816631153416877)
        (9.8, 0.81737104055273)
        (9.9, 0.817566081521138)
        (10.0, 0.818234103840704)
        (10.1, 0.81877251049493)
        (10.2, 0.819336040772037)
        (10.3, 0.819454104526464)
        (10.4, 0.819400090358813)
        (10.5, 0.819369767438769)
        };
    \legend{SVE, SR, ENT}
     
    \end{axis}
    \end{tikzpicture}
    }
}
\caption{UCI-121 test set accuracy for a given train step (in thousands). SVE - shared embedding method, ENT - $1.05$-entmax with embeddings initialized from $\mathcal{N}(0, 1)$, SR - stable rank with $\alpha_{\text{sr}} = 0.05$.}
\label{fig:train-test-acc}
\end{center}
\vskip -0.2in
\end{figure}
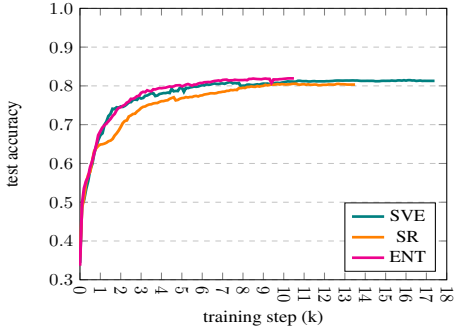

Details of the training process are displayed in Figure \ref{fig:train-test-acc}. Both the shared embedding method and the $1.05$-entmax method show similar characteristics through the earlier part of the training process. A major difference, however, is that the entmax method hits the stop criterion significantly earlier and provides a higher final test set accuracy. The stable rank method with $\alpha_{\text{sr}} = 0.05$ hits visibly lower accuracy levels throughout training, while requiring more steps than the $1.05$-entmax method. Figure \ref{fig:train-steps-sr-train-steps} directly compares the number of steps before reaching the maximum test set accuracy. In particular, the $1.05$-entmax model reaches its peak test set accuracy after $10\,300$ steps compared to $16\,200$ for the shared embedding method, which is a $36.4\%$ decrease in training time measured in steps. The stable rank model with $\alpha_{\text{sr}} = 0.05$ requires $10\,600$ steps.

Ablation studies for methods involving orthogonalization, stable rank and von Neumann entropy as means to enforce independence in the shared embeddings are presented in Table \ref{tab:accuracy-orth-sr-vn}. These results suggest that orthogonalization and, in particular, von Neumann entropy have an adverse effect on the final classification accuracy, while the stable rank restrictions do not seem to improve the results relative to straightforward shared embeddings but they do not decisively hurt them either. An extensive ablation for the sparse attention methods is presented in Table \ref{tab:accuracy-sparse-attention}. The $\alpha$-entmax approaches are evaluated in two distinct settings. In the first one, $\alpha$ is picked as a hyperparameter, constant across the whole training procedure. In the second one, $\alpha$ is treated as a model parameter, with an initial value, and is optimized with gradient descent. In this case, the final optimized value of $\alpha$ is reported. The evaluations of the $\alpha$-entmax methods show that it is important to adjust the weight initialization procedure of the model. For embeddings initialized as in the vanilla variable embedding approach, the final accuracy suffers. Increasing the standard deviation of the normal distribution from which the initial weights are sampled markedly improves the test accuracy. In particular, for the $1.05$-entmax method with a standard normal distribution used for initialization, the accuracy reaches levels higher than for any other considered setup, with a significantly shortened training time. This suggests that for the particular tackled set of tasks, an attention mechanism with moderate induced sparsity provides slight advantages in terms of accuracy and significant advantages in terms of training time relative to the standard attention mechanism.

\subsection{Interpretability}
\label{sec:interpretability}
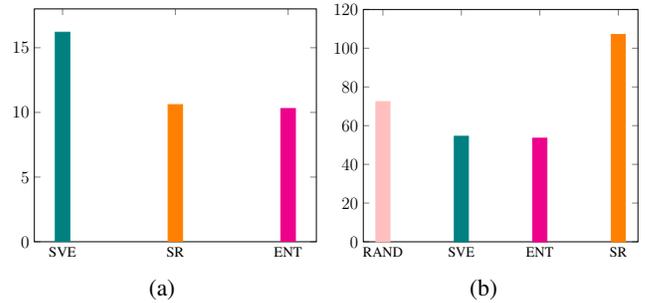
\begin{figure}
  \begin{subfigure}{0.5\columnwidth}
    \resizebox{\columnwidth}{0.8\columnwidth}{
      \begin{tikzpicture}
        \begin{axis}[
            y tick label style={
                /pgf/number format/.cd,
                fixed,
                fixed zerofill,
                precision=0,
                /tikz/.cd,
                font=\Large
            },
            xmin=0.75,
            xmax=3.25,
            ymin=0,
            ymax=18,
            xtick={1,2,3},
            xticklabels={SVE,SR,ENT},
            every axis plot/.append style={
              ybar,
              bar width=10,
              bar shift=0pt,
              fill
            }
          ]
          \addplot[teal]coordinates {(1,16.2)};
          \addplot[orange]coordinates{(2,10.6)};
          \addplot[magenta]coordinates{(3,10.3)};
        \end{axis}
      \end{tikzpicture}
    }
    \caption{} \label{fig:train-steps-sr-train-steps}
  \end{subfigure}%
  \hspace*{\fill}
  \begin{subfigure}{0.5\columnwidth}
    \resizebox{\columnwidth}{0.83\columnwidth}{
      \begin{tikzpicture}
        \begin{axis}[
            y tick label style={
                /pgf/number format/.cd,
                fixed,
                fixed zerofill,
                precision=0,
                /tikz/.cd,
                font=\Large
            },
            xmin=0.75,
            xmax=4.25,
            ymin=0,
            ymax=120,
            xtick={1,2,3,4},
            xticklabels={RAND, SVE, ENT, SR},
            every axis plot/.append style={
              ybar,
              bar width=10,
              bar shift=0pt,
              fill
            }
          ]
          \addplot[pink]coordinates {(1,72.4)};
          \addplot[teal]coordinates {(2,54.5)};
          \addplot[magenta]coordinates{(3,53.5)};
          \addplot[orange]coordinates{(4,107.1)};
        \end{axis}
      \end{tikzpicture}
    }
    \caption{} \label{fig:train-steps-sr-sr}
  \end{subfigure}%
\caption{(a) Training steps to reach best test set accuracy. (b) Stable rank of the shared embedding matrix after training - best accuracy model. SVE - shared embedding method, ENT - $1.05$-entmax with embeddings initialized from $\mathcal{N}(0, 1)$, SR - stable rank with $\alpha_{\text{sr}} = 0.05$, RAND - random embedding matrix with entries from $\mathcal{N}(0, 1)$.} \label{fig:train-steps-sr}
\end{figure}

To verify whether the introduction of shared embeddings $\{\mathbf{s}_{k}\}_{k = 1}^{D}$ in fact generates reusable concepts and provides a degree of interpretability, we investigate their characteristics. The degree to which the shared embeddings are independent after training, as measured by stable rank, is presented in Figure \ref{fig:train-steps-sr-sr}. The embeddings obtained from the straightforward shared embedding method and its version with sparse attention seem significantly less independent than random embeddings. On the flip side, the stable rank incarnation of our method is able to visibly increase the independence of the embeddings, as it directly optimizes for this goal.

We proceed to investigate whether this notion of independence correlates with the interpretability of specific shared embeddings. Towards this end, we propose an evaluation protocol for the ability of shared variable embeddings to differentiate between real-world concepts:
\begin{itemize}
    \item Compute the attention scores for all raw variable embeddings.
    \item Sample $\mathbf{s}_{p}$ without repetition from the set of shared embeddings $\{\mathbf{s}_{k}\}_{k = 1}^{D}$.
    \item Select $K$ variables from the given tasks whose raw variable embeddings are most similar to $\mathbf{s}_{p}$.
    \item Verify whether the selected $K$ variables map intuitively to one or more real-world concepts.
\end{itemize}
The variables most similar to a shared embedding representing a specific concept would be expected to share some intuitive notion, category or semantic meaning. In our evaluations, we choose $K = 5$. To aid in a quantitative as well as qualitative assessment of mapping to concepts, we introduce a measurable and less subjective assignment to real-world concepts in the form of the Subject Area ascribed to a given task/dataset in the UCI repository \cite{Kelly2023}. Each dataset has one Subject Area (SA) assigned to it from the following $11$ possibilities: Biology (Bio), Business (Bus), Climate and Environment (C\&E), Computer Science (CS), Engineering (E), Games (G), Health and Medicine (H\&M), Law (L), Physics and Chemistry (P\&C), Social Sciences (SS) and Other (O). These categories afford us the option to measure the performance of a given model in terms of interpretability and compare it with other methods. 
\begin{table}
\caption{Classification accuracy for variable embedding methods with $\alpha$-entmax sparse attention on the UCI-121 test set. $\alpha$ represents the initial value used, while OPTIMIZED $\alpha$ is the final value of $\alpha$ for methods where $\alpha$ is treated as a model parameter.}
\label{tab:accuracy-sparse-attention}
\vskip 0.15in
\begin{center}
\begin{small}
\begin{sc}
\begin{tabular}{lccc}
\toprule
$\alpha$ & Context std & Accuracy & Optimized $\alpha$ \\
\midrule
0.5 & 1.0 & 79.7 & 0.91 \\
0.9 & 1.0 & 81.1 & 0.83 \\
1.05 & 0.01 & 80.9 & 1.74 \\
1.05 & 0.05 & 78.8 & 1.58 \\
1.05 & 0.1 & 80.9 & 1.62 \\
1.05 & 0.5 & 81.1 & 1.36 \\
\textbf{1.05} & \textbf{1.0} & \textbf{81.9} & \textbf{1.00} \\
1.05 & 2.0 & 76.0 & 0.88 \\
1.5 & 0.1 & 78.9 & 1.76 \\
1.5 & 0.5 & 80.8 & $\times$ \\
1.5 & 1.0 & 77.0 & $\times$ \\
1.5 & 1.0 & 78.6 & 1.36 \\
1.5 & 1.0 & 80.8 & 1.38 \\
2.0 & 1.0 & 75.2 & $\times$ \\
\bottomrule
\end{tabular}
\end{sc}
\end{small}
\end{center}
\vskip -0.1in
\end{table}

Table \ref{tab:interpretability-shared-embeddings-one} and Table \ref{tab:interpretability-sparse-attention-one} show the results of running our evaluation procedure for one randomly chosen shared embedding. This consists of $5$ variables most similar to the sampled shared embedding. We also present extended results where this procedure is repeated in Appendix \ref{app:interpretability} and results on SVEs commonly shared by tasks in Appendix \ref{app:sves-shared-commonly}.

Table \ref{tab:interpretability-shared-embeddings-one} suggests that the shared embedding method generates an embedding which is most similar to variables which have an intuitive interpretation of measuring physical quantities and phenomena. A quantitative analysis confirms this qualitative assessment. All identified variables belong to the Physics and Chemistry Subject Area. Also, all the variables come from distinct datasets. Qualitatively, these most similar variables represent quantities related to physical processes, e.g. energies, waveforms, as well as objects which such quantities describe: molecules, particles etc. They do seem to carry with them a distinct intuitive meaning. Both the quantitative and qualitative results indicate that the proposed method is able to identify concepts rather than tie the embeddings to specific datasets, contrary to what is the case for the standard variable embedding approach.
\begin{table}
\caption{Most similar variables for a random choice of a shared embedding. Shared embedding method. Variables sorted in descending order of similarity. 
(-) denotes ambiguous data.}
\label{tab:interpretability-shared-embeddings-one}
\vskip 0.15in
\begin{center}
\begin{small}
\begin{sc}
\begin{tabular}{p{0.2\linewidth} p{0.5\linewidth} p{0.12\linewidth}}
\toprule
Dataset & Variable meaning & SA \\
\midrule
MUSK (V2) & \textit{A distance feature of a molecule along a ray.} & P\&C \\
\greyrule
Conn. Bench (S,MvR) & \textit{Energy within a particular frequency band, integrated over a certain period of time.} & P\&C \\
\greyrule
Waveform (V1) & \textit{Waveform feature; contains noise but is not all noise.} & P\&C \\
\greyrule
MiniBooNE & \textit{A particle ID variable (real) for an event.} & P\&C \\
\greyrule
Annealing & - & P\&C \\
\bottomrule
\end{tabular}
\end{sc}
\end{small}
\end{center}
\vskip -0.1in
\end{table}

Table \ref{tab:interpretability-sparse-attention-one} shows a similar result for the $1.05$-entmax sparse attention method. In this case, the majority of the selected variables seem to represent concepts related to health or biological systems. In quantitative terms, the majority belongs to one Subject Area - Health and Medicine, while there is also one variable identified as coming from the Biology Subject Area. The one variable which does not fit the health or biological interpretation is the least similar from the selected $5$ and it belongs to the by-definition-broad Other category. Qualitatively, the most similar variables relate to living organisms. This is also the case for the fourth most similar variable from the Biology SA. We do, however, notice less internal consistency in this grouping, relative to the results from the base SVE method. Both quantitative and qualitative results suggest that the base shared embedding method may actually produce more interpretable shared embeddings than the sparse attention approach. This is supported by repeat analysis presented in Appendix \ref{app:interpretability} where both methods seem to produce interpretable embeddings but the base method outperforms the sparse attention method in terms of the cohesion of the embeddings. The base method produces representations which are more easily linked to one broad intuitive concept while the inclusion of sparse attention prefers embeddings which are linked to more than one but still related concepts (e.g. biological and health-related ones).
\begin{table}
\caption{Most similar variables for a random choice of a shared embedding. $1.05$-entmax sparse attention method with embeddings initialized from $\mathcal{N}(0, 1)$. Variables sorted in descending order of similarity. (-) denotes ambiguous data.}
\label{tab:interpretability-sparse-attention-one}
\vskip 0.15in
\begin{center}
\begin{small}
\begin{sc}
\begin{tabular}{p{0.2\linewidth} p{0.5\linewidth} p{0.12\linewidth}}
\toprule
Dataset & Variable meaning & SA \\
\midrule
Breast Cancer WI (D.) & \textit{Mean compactness of the cell nuclei in the image.} & H\&M \\
\greyrule
Thyroid Disease & - & H\&M \\
\greyrule
Arrhythmia & - & H\&M \\
\greyrule
Leaves (Shape) & \textit{A specific feature relating to the shape of the leaf.} & Bio \\
\greyrule
Synth. Control & \textit{Point value on synthetically generated control chart.} & O \\
\bottomrule
\end{tabular}
\end{sc}
\end{small}
\end{center}
\vskip -0.1in
\end{table}

While this investigation points to the relative performance of the methods, it is worth analyzing whether the results are not merely caused by the statistical characteristics of the dataset. To facilitate this, we replace the evaluation procedure with its random counterpart where we sample $K = 5$ variables from the UCI-121 dataset and perform the same qualitative and quantitative assessment as was the case for variables similar to the shared embedding from our models. The results for one such sample are presented in Table \ref{tab:interpretability-random-one}. There is one repeated category (H\&M), however, it does not have a majority. Also, other than in some samples for the $1.05$-entmax method, there is only one repeated category, not a contest between two categories. These indications also hold across additional samples presented in Appendix \ref{app:interpretability}. A further differentiating characteristic is that all the variables within each of the extended samples (in Appendix \ref{app:interpretability}) come from different datasets, whereas for our analyzed models datasets occasionally repeat. The only case when we observe a majority category for the random variables is one where these variables represent the Other SA, which is, by definition, a broad bracket in which we do not expect the variables to represent similar concepts. All this points to the fact that the most similar variables obtained from out models are significantly different from random choice.

\subsection{Accuracy vs. interpretability trade-off}
\label{sec:trade-off}
Drawing on the quantitative and qualitative results, we find that for our best performing methods none of them strictly dominates the other in terms of \textit{both} accuracy and interpretability. The $1.05$-entmax method achieves higher final accuracy than our base shared embedding method, $81.9\%$ vs. $81.5\%$. On the flip side, the base shared embedding method does seem to more successfully separate real-world concepts into specific shared embeddings. 
\begin{table}
\caption{Random choice of variables from the UCI-121 dataset. (-) denotes ambiguous data, (*) denotes inferred Subject Areas.}
\label{tab:interpretability-random-one}
\vskip 0.15in
\begin{center}
\begin{small}
\begin{sc}
\begin{tabular}{p{0.2\linewidth} p{0.5\linewidth} p{0.12\linewidth}}
\toprule
Dataset & Variable meaning & SA \\
\midrule
Audiology (S.) & - & H\&M \\
\greyrule
- & - & Bio(*) \\
\greyrule
Synth. Control & \textit{Point value on synthetically generated control chart.} & O \\
\greyrule
Tic-Tac-Toe End. & \textit{State of the bottom-left square at the end of a game.} & G \\
\greyrule
Heart Dis. & - & H\&M \\
\bottomrule
\end{tabular}
\end{sc}
\end{small}
\end{center}
\vskip -0.1in
\end{table}
Specifically, in our extended results (Appendix \ref{app:interpretability}), we see that the base shared embedding method achieves more consistent concept assignment to the shared embeddings. For $5$ trials, we obtain $3$ where there is a majority SA. In one trial, there was no majority but there was a dominant SA without draws. One trial resulted in a draw between SAs. It is also worth noting that for one trial, all the similar variables come from the same SA (P\&C) and from different datasets. Conversely, for the $1.05$-entmax method, the assignment of shared embeddings to concepts is still present, only weaker. One trial results in a majority SA assignment (H\&M). $3$ trials end in draws between dominant categories. Importantly, one trial has all the similar variables represent different SAs from distinct datasets. With these results, there seems to be a trade-off between prediction accuracy and interpretability. It should be noted that, even in the presence of such a trade-off, both the base shared embedding method and the sparse attention method more decisively than random choice link specific shared embeddings to intuitive concepts. A discussion on this is presented in Appendix \ref{app:interpretability-random}.

\section{Conclusion}
We have proposed a new variable embedding architecture for general prediction problems. This architecture is based on shared embeddings with attention, which is a lightweight addition to the variable embedding architecture. We have considered several potential versions of this approach, introducing restrictions on the shared embeddings and adding sparsity to the attention mechanism. Other than in the standard variable embedding method, our approach does not require one variable embedding to represent one specific variable from a concrete dataset, but rather encourages the reuse of shared embeddings among variables across distinct datasets.

In empirical experiments, we have shown that our base method performs as well as the standard variable embedding method on the UCI-121 dataset, while not requiring any fine-tuning, which the standard method does. Additionally, we have performed a series of ablations to identify which versions of our architecture perform favorably in terms of classification accuracy and the potential interpretability of the shared embeddings. The results have demonstrated that the sparse attention mechanism helps in: (1) achieving superior classification performance and (2) requiring significantly less training steps than our base SVE method. However, the gain comes at a cost of decreased interpretability relative to our base shared embedding method. This suggests a potential trade-off between performance and interpretability.

As far as interpretability itself is concerned, both our base method and its extension with sparse attention are able to use the shared embeddings to identify abstract concepts instead of making hard links to concrete variables from specific datasets, which is the case for the standard variable embedding approach. The base shared embedding method generates embeddings which are more interpretable and internally consistent than the sparse attention modification.

Given the results we have obtained, several lines of enquiry emerge: (1) investigation of other methods to restrict the shared embedding space, e.g. based on quantization, (2) adaptation of the variable embedding method and the shared embedding approach to vision, (3) use of \textit{self-supervised learning} for the shared embeddings approach.

\section*{Acknowledgements}
This research was carried out with the support of the Laboratory of Bioinformatics and Computational Genomics and the High Performance Computing Center of the Faculty of Mathematics and Information Science Warsaw University of Technology.

\bibliography{aaai25}

\begin{thebibliography}{44}
\providecommand{\natexlab}[1]{#1}

\bibitem[{Assran et~al.(2023)Assran, Duval, Misra, Bojanowski, Vincent, Rabbat, LeCun, and Ballas}]{Assran2023}
Assran, M.; Duval, Q.; Misra, I.; Bojanowski, P.; Vincent, P.; Rabbat, M.; LeCun, Y.; and Ballas, N. 2023.
\newblock Self-Supervised Learning From Images With a Joint-Embedding Predictive Architecture.
\newblock In \emph{Proceedings of the IEEE/CVF Conference on Computer Vision and Pattern Recognition (CVPR)}, 15619--15629.

\bibitem[{Bahdanau, Cho, and Bengio(2015)}]{Bahdanau2015}
Bahdanau, D.; Cho, K.; and Bengio, Y. 2015.
\newblock Neural Machine Translation by Jointly Learning to Align and Translate.
\newblock In Bengio, Y.; and LeCun, Y., eds., \emph{3rd International Conference on Learning Representations, {ICLR} 2015, San Diego, CA, USA, May 7-9, 2015, Conference Track Proceedings}.

\bibitem[{Bardes, Ponce, and LeCun(2022)}]{Bardes2022}
Bardes, A.; Ponce, J.; and LeCun, Y. 2022.
\newblock {VICR}eg: Variance-Invariance-Covariance Regularization for Self-Supervised Learning.
\newblock In \emph{International Conference on Learning Representations}.

\bibitem[{Bengio, Ducharme, and Vincent(2000)}]{Bengio2000}
Bengio, Y.; Ducharme, R.; and Vincent, P. 2000.
\newblock A Neural Probabilistic Language Model.
\newblock In Leen, T.~K.; Dietterich, T.~G.; and Tresp, V., eds., \emph{NIPS}, 932--938. MIT Press.

\bibitem[{Br\"uggemann et~al.(2021)Br\"uggemann, Kanakis, Obukhov, Georgoulis, and Van~Gool}]{Bruggemann2021}
Br\"uggemann, D.; Kanakis, M.; Obukhov, A.; Georgoulis, S.; and Van~Gool, L. 2021.
\newblock Exploring Relational Context for Multi-Task Dense Prediction.
\newblock In \emph{Proceedings of the IEEE/CVF International Conference on Computer Vision (ICCV)}, 15869--15878.

\bibitem[{Caruana(1993)}]{Caruana1993}
Caruana, R. 1993.
\newblock Multitask Learning: A Knowledge-Based Source of Inductive Bias.
\newblock In \emph{Proceedings of the Tenth International Conference on International Conference on Machine Learning}, ICML'93, 41–48. San Francisco, CA, USA: Morgan Kaufmann Publishers Inc.
\newblock ISBN 1558603077.

\bibitem[{Caruana(1994)}]{Caruana1994}
Caruana, R. 1994.
\newblock Learning Many Related Tasks at the Same Time with Backpropagation.
\newblock In Tesauro, G.; Touretzky, D.; and Leen, T., eds., \emph{Advances in Neural Information Processing Systems}, volume~7. MIT Press.

\bibitem[{Caruana(1996)}]{Caruana1996}
Caruana, R. 1996.
\newblock Algorithms and Applications for Multitask Learning.
\newblock In \emph{International Conference on Machine Learning}.

\bibitem[{Caruana(1997)}]{Caruana1997}
Caruana, R. 1997.
\newblock Multitask Learning.
\newblock \emph{Machine Learning}, 28: 41--75.

\bibitem[{Chen et~al.(2020)Chen, Kornblith, Norouzi, and Hinton}]{Chen2020}
Chen, T.; Kornblith, S.; Norouzi, M.; and Hinton, G. 2020.
\newblock A Simple Framework for Contrastive Learning of Visual Representations.
\newblock In III, H.~D.; and Singh, A., eds., \emph{Proceedings of the 37th International Conference on Machine Learning}, volume 119 of \emph{Proceedings of Machine Learning Research}, 1597--1607. PMLR.

\bibitem[{Cui et~al.(2021)Cui, Qi, Gu, You, Zhang, and Harada}]{Cui2021}
Cui, Z.; Qi, G.-J.; Gu, L.; You, S.; Zhang, Z.; and Harada, T. 2021.
\newblock Multitask AET With Orthogonal Tangent Regularity for Dark Object Detection.
\newblock In \emph{Proceedings of the IEEE/CVF International Conference on Computer Vision (ICCV)}, 2553--2562.

\bibitem[{Fern{{\'a}}ndez-Delgado et~al.(2014)Fern{{\'a}}ndez-Delgado, Cernadas, Barro, and Amorim}]{Delgado2014}
Fern{{\'a}}ndez-Delgado, M.; Cernadas, E.; Barro, S.; and Amorim, D. 2014.
\newblock Do we Need Hundreds of Classifiers to Solve Real World Classification Problems?
\newblock \emph{Journal of Machine Learning Research}, 15(90): 3133--3181.

\bibitem[{Fukushima(1980)}]{Fukushima1980}
Fukushima, K. 1980.
\newblock {N}eocognitron: {A} Self-Organizing Neural Network Model for a Mechanism of Pattern Recognition Unaffected by Shift in Position.
\newblock \emph{Biological Cybernetics}, 36: 193--202.

\bibitem[{Gao et~al.(2019)Gao, Ma, Zhao, Liu, and Yuille}]{Gao2019}
Gao, Y.; Ma, J.; Zhao, M.; Liu, W.; and Yuille, A.~L. 2019.
\newblock NDDR-CNN: Layerwise Feature Fusing in Multi-Task CNNs by Neural Discriminative Dimensionality Reduction.
\newblock In \emph{Proceedings of the IEEE/CVF Conference on Computer Vision and Pattern Recognition (CVPR)}.

\bibitem[{Goyal et~al.(2021)Goyal, Lamb, Hoffmann, Sodhani, Levine, Bengio, and Sch{\"o}lkopf}]{Goyal2021}
Goyal, A.; Lamb, A.; Hoffmann, J.; Sodhani, S.; Levine, S.; Bengio, Y.; and Sch{\"o}lkopf, B. 2021.
\newblock Recurrent Independent Mechanisms.
\newblock In \emph{International Conference on Learning Representations}.

\bibitem[{Graves, Wayne, and Danihelka(2014)}]{Graves2014}
Graves, A.; Wayne, G.; and Danihelka, I. 2014.
\newblock Neural Turing Machines.
\newblock \emph{arXiv:1410.5401}.

\bibitem[{Graves et~al.(2016)Graves, Wayne, Reynolds, Harley, Danihelka, Grabska-Barwińska, Colmenarejo, Grefenstette, Ramalho, Agapiou, Badia, Hermann, Zwols, Ostrovski, Cain, King, Summerfield, Blunsom, Kavukcuoglu, and Hassabis}]{Graves2016}
Graves, A.; Wayne, G.; Reynolds, M.; Harley, T.; Danihelka, I.; Grabska-Barwińska, A.; Colmenarejo, S.~G.; Grefenstette, E.; Ramalho, T.; Agapiou, J.; Badia, A.~P.; Hermann, K.~M.; Zwols, Y.; Ostrovski, G.; Cain, A.; King, H.; Summerfield, C.; Blunsom, P.; Kavukcuoglu, K.; and Hassabis, D. 2016.
\newblock Hybrid computing using a neural network with dynamic external memory.
\newblock \emph{Nature}, 538(7626): 471--476.

\bibitem[{Hinton and Roweis(2002)}]{Hinton2002}
Hinton, G.~E.; and Roweis, S. 2002.
\newblock Stochastic Neighbor Embedding.
\newblock In Becker, S.; Thrun, S.; and Obermayer, K., eds., \emph{Advances in Neural Information Processing Systems}, volume~15. MIT Press.

\bibitem[{Hu and Singh(2021)}]{Hu2021}
Hu, R.; and Singh, A. 2021.
\newblock UniT: Multimodal Multitask Learning With a Unified Transformer.
\newblock In \emph{Proceedings of the IEEE/CVF International Conference on Computer Vision (ICCV)}, 1439--1449.

\bibitem[{Kaiser et~al.(2017)Kaiser, Gomez, Shazeer, Vaswani, Parmar, Jones, and Uszkoreit}]{Kaiser2017}
Kaiser, L.; Gomez, A.~N.; Shazeer, N.; Vaswani, A.; Parmar, N.; Jones, L.; and Uszkoreit, J. 2017.
\newblock One Model To Learn Them All.
\newblock \emph{arXiv:1706.05137}.

\bibitem[{Kelly, Longjohn, and Nottingham(2023)}]{Kelly2023}
Kelly, M.; Longjohn, R.; and Nottingham, K. 2023.
\newblock The UCI Machine Learning Repository.

\bibitem[{LeCun et~al.(1989)LeCun, Boser, Denker, Henderson, Howard, Hubbard, and Jackel}]{LeCun1989}
LeCun, Y.; Boser, B.; Denker, J.~S.; Henderson, D.; Howard, R.~E.; Hubbard, W.; and Jackel, L.~D. 1989.
\newblock Backpropagation Applied to Handwritten Zip Code Recognition.
\newblock \emph{Neural Computation}, 1: 541--551.

\bibitem[{Mahmud and Ray(2007)}]{Mahmud2007}
Mahmud, M.; and Ray, S. 2007.
\newblock Transfer Learning using Kolmogorov Complexity: Basic Theory and Empirical Evaluations.
\newblock In Platt, J.; Koller, D.; Singer, Y.; and Roweis, S., eds., \emph{Advances in Neural Information Processing Systems}, volume~20. Curran Associates, Inc.

\bibitem[{Martins and Astudillo(2016)}]{Martins2016}
Martins, A. F.~T.; and Astudillo, R.~F. 2016.
\newblock From Softmax to Sparsemax: A Sparse Model of Attention and Multi-Label Classification.
\newblock In \emph{Proceedings of the 33rd International Conference on International Conference on Machine Learning - Volume 48}, ICML'16, 1614–1623. JMLR.org.

\bibitem[{Mcculloch and Pitts(1943)}]{McCulloch1943}
Mcculloch, W.; and Pitts, W. 1943.
\newblock A Logical Calculus of Ideas Immanent in Nervous Activity.
\newblock \emph{Bulletin of Mathematical Biophysics}, 5: 127--147.

\bibitem[{{McInnes}, {Healy}, and {Melville}(2018)}]{McInnes2018}
{McInnes}, L.; {Healy}, J.; and {Melville}, J. 2018.
\newblock UMAP: Uniform Manifold Approximation and Projection for Dimension Reduction.
\newblock \emph{arXiv:1802.03426}.

\bibitem[{Meyerson and Miikkulainen(2019)}]{Meyerson2019}
Meyerson, E.; and Miikkulainen, R. 2019.
\newblock Modular Universal Reparameterization: Deep Multi-task Learning Across Diverse Domains.
\newblock In Wallach, H.; Larochelle, H.; Beygelzimer, A.; d\textquotesingle Alch\'{e}-Buc, F.; Fox, E.; and Garnett, R., eds., \emph{Advances in Neural Information Processing Systems}, volume~32. Curran Associates, Inc.

\bibitem[{Meyerson and Miikkulainen(2021)}]{Meyerson2021}
Meyerson, E.; and Miikkulainen, R. 2021.
\newblock The Traveling Observer Model: Multi-task Learning Through Spatial Variable Embeddings.
\newblock In \emph{International Conference on Learning Representations}.

\bibitem[{Mikolov et~al.(2013)Mikolov, Chen, Corrado, and Dean}]{Mikolov2013}
Mikolov, T.; Chen, K.; Corrado, G.; and Dean, J. 2013.
\newblock Efficient Estimation of Word Representations in Vector Space.
\newblock In Bengio, Y.; and LeCun, Y., eds., \emph{1st International Conference on Learning Representations, {ICLR} 2013, Scottsdale, Arizona, USA, May 2-4, 2013, Workshop Track Proceedings}.

\bibitem[{Misra et~al.(2016)Misra, Shrivastava, Gupta, and Hebert}]{Misra2016}
Misra, I.; Shrivastava, A.; Gupta, A.; and Hebert, M. 2016.
\newblock Cross-Stitch Networks for Multi-Task Learning.
\newblock In \emph{Proceedings of the IEEE Conference on Computer Vision and Pattern Recognition (CVPR)}.

\bibitem[{Mnih et~al.(2015)Mnih, Kavukcuoglu, Silver, Rusu, Veness, Bellemare, Graves, Riedmiller, Fidjeland, Ostrovski, Petersen, Beattie, Sadik, Antonoglou, King, Kumaran, Wierstra, Legg, and Hassabis}]{Mnih2015}
Mnih, V.; Kavukcuoglu, K.; Silver, D.; Rusu, A.~A.; Veness, J.; Bellemare, M.~G.; Graves, A.; Riedmiller, M.; Fidjeland, A.~K.; Ostrovski, G.; Petersen, S.; Beattie, C.; Sadik, A.; Antonoglou, I.; King, H.; Kumaran, D.; Wierstra, D.; Legg, S.; and Hassabis, D. 2015.
\newblock Human-level control through deep reinforcement learning.
\newblock \emph{Nature}, 518(7540): 529--533.

\bibitem[{Perez et~al.(2018)Perez, Strub, de~Vries, Dumoulin, and Courville}]{Perez2018}
Perez, E.; Strub, F.; de~Vries, H.; Dumoulin, V.; and Courville, A.~C. 2018.
\newblock FiLM: Visual Reasoning with a General Conditioning Layer.
\newblock In \emph{AAAI}.

\bibitem[{Peters, Niculae, and Martins(2019)}]{Peters2019}
Peters, B.; Niculae, V.; and Martins, A. F.~T. 2019.
\newblock Sparse Sequence-to-Sequence Models.
\newblock In Korhonen, A.; Traum, D.; and M{\`a}rquez, L., eds., \emph{Proceedings of the 57th Annual Meeting of the Association for Computational Linguistics}, 1504--1519. Florence, Italy: Association for Computational Linguistics.

\bibitem[{Rombach et~al.(2022)Rombach, Blattmann, Lorenz, Esser, and Ommer}]{Rombach2022}
Rombach, R.; Blattmann, A.; Lorenz, D.; Esser, P.; and Ommer, B. 2022.
\newblock High-Resolution Image Synthesis With Latent Diffusion Models.
\newblock In \emph{Proceedings of the IEEE/CVF Conference on Computer Vision and Pattern Recognition (CVPR)}, 10684--10695.

\bibitem[{Rosenblatt(1958)}]{Rosenblatt1958}
Rosenblatt, F. 1958.
\newblock {The perceptron: A probabilistic model for information storage and organization in the brain.}
\newblock \emph{Psychological Review}, 65(6): 386--408.

\bibitem[{Thrun and O'Sullivan(1996)}]{Thrun1996}
Thrun, S.; and O'Sullivan, J. 1996.
\newblock Discovering Structure in Multiple Learning Tasks: The {TC} Algorithm.
\newblock In Saitta, L., ed., \emph{Proceedings of the 13th International Conference on Machine Learning ICML-96}. San Mateo, CA: Morgen Kaufmann.

\bibitem[{Tsallis(1988)}]{Tsallis1988}
Tsallis, C. 1988.
\newblock Possible generalization of {Boltzmann-Gibbs} statistics.
\newblock \emph{Journal of Statistical Physics}, 52: 479--487.

\bibitem[{van~den Oord, Vinyals, and Kavukcuoglu(2017)}]{vandenOord2017}
van~den Oord, A.; Vinyals, O.; and Kavukcuoglu, K. 2017.
\newblock Neural Discrete Representation Learning.
\newblock In Guyon, I.; Luxburg, U.~V.; Bengio, S.; Wallach, H.; Fergus, R.; Vishwanathan, S.; and Garnett, R., eds., \emph{Advances in Neural Information Processing Systems}, volume~30. Curran Associates, Inc.

\bibitem[{Vasershtein(1971)}]{Vasershtein1971}
Vasershtein, L. 1971.
\newblock Stable rank of rings and dimensionality of topological spaces.
\newblock \emph{Functional Analysis and Its Applications}, 5: 102--110.

\bibitem[{Vaswani et~al.(2017)Vaswani, Shazeer, Parmar, Uszkoreit, Jones, Gomez, Kaiser, and Polosukhin}]{Vaswani2017}
Vaswani, A.; Shazeer, N.; Parmar, N.; Uszkoreit, J.; Jones, L.; Gomez, A.~N.; Kaiser, L.~u.; and Polosukhin, I. 2017.
\newblock Attention is All you Need.
\newblock In Guyon, I.; Luxburg, U.~V.; Bengio, S.; Wallach, H.; Fergus, R.; Vishwanathan, S.; and Garnett, R., eds., \emph{Advances in Neural Information Processing Systems}, volume~30. Curran Associates, Inc.

\bibitem[{Yang and Hospedales(2014)}]{Yang2014}
Yang, Y.; and Hospedales, T.~M. 2014.
\newblock A Unified Perspective on Multi-Domain and Multi-Task Learning.
\newblock In \emph{ICLR}.

\bibitem[{Ye and Xu(2023)}]{Ye2023}
Ye, H.; and Xu, D. 2023.
\newblock TaskPrompter: Spatial-Channel Multi-Task Prompting for Dense Scene Understanding.
\newblock In \emph{The Eleventh International Conference on Learning Representations}.

\bibitem[{Zbontar et~al.(2021)Zbontar, Jing, Misra, LeCun, and Deny}]{Zbontar2021}
Zbontar, J.; Jing, L.; Misra, I.; LeCun, Y.; and Deny, S. 2021.
\newblock Barlow Twins: Self-Supervised Learning via Redundancy Reduction.
\newblock In Meila, M.; and Zhang, T., eds., \emph{Proceedings of the 38th International Conference on Machine Learning}, volume 139 of \emph{Proceedings of Machine Learning Research}, 12310--12320. PMLR.

\bibitem[{Zintgraf et~al.(2019)Zintgraf, Shiarli, Kurin, Hofmann, and Whiteson}]{Zintgraf2019}
Zintgraf, L.; Shiarli, K.; Kurin, V.; Hofmann, K.; and Whiteson, S. 2019.
\newblock Fast Context Adaptation via Meta-Learning.
\newblock In Chaudhuri, K.; and Salakhutdinov, R., eds., \emph{Proceedings of the 36th International Conference on Machine Learning}, volume~97 of \emph{Proceedings of Machine Learning Research}, 7693--7702. PMLR.

\end{thebibliography}

\onecolumn
\appendix

\section{Details of methods imposing independence of shared variable embeddings}
\label{app:restrictions}
\subsection{Orthogonalization}
For a simple notion of independence, we consider an embedding to be independent from other embeddings when it is not a linear combination of them. With this, $r_{\mathbf{S}} = \text{rank}(\mathbf{S})$ is a measure of independence. In a realistic setting, we might still have $r_{\mathbf{S}} = C$ even if multiple embeddings are approximately linearly dependent. An operational measure of the rank of $\mathbf{S}$ would require to address this drawback and we describe such a measure $q_{\mathbf{S}}$ in Section \ref{sec:stable-rank}. The results from Section \ref{sec:interpretability} show that the proposed training procedure results in $q_{\mathbf{S}} < C$. A straightforward way to build in more independence is to require $\mathbf{S}$ to consist of \textit{orthonormal} vectors. With an additional assumption of $D = C$, this would translate into an \textit{orthogonality} requirement, which could be incorporated in the loss function:
\begin{equation}
    L_{\text{orth}}(\hat{\mathbf{y}}, \mathbf{t}) = L(\hat{\mathbf{y}}, \mathbf{t}) + \alpha_{\text{orth}}\left(\sum_{i = j}{\left(1 - \mathbf{D}_{i, j}\right)^{2}} + \sum_{i \neq j}{\mathbf{D}_{i, j}^{2}}\right)
\end{equation}
where $\mathbf{D}_{C \times C} = \mathbf{S}^{T}\mathbf{S} = \mathbf{I}$. A subtle problem is that, for random initializations of $\mathbf{S}$, we might have $\det{\left(\mathbf{S}^{T}\mathbf{S}\right)} = -1$ and the optimization procedure may have trouble updating $\mathbf{S}$ to obtain $\mathbf{S}^{T}\mathbf{S} \approx \mathbf{I}$. Because of that, the weight initialization procedure has to be adjusted so that $\det{\left(\mathbf{S}^{T}\mathbf{S}\right)} = 1$. This is done by only allowing random initializations which result in $\det{\left(\mathbf{S}^{T}\mathbf{S}\right) = 1}$.

\subsection{Stable rank}
\label{sec:stable-rank}
Instead of focusing on restricting $\mathbf{S}$, it is possible to explicitly add $r_{\mathbf{S}}$ to the loss function. A significant drawback of this is the discontinuous characteristic of the rank measure, which makes it unsuitable for gradient descent. To address this, we rely on a continuous proxy. Let us consider a matrix $\mathbf{A}_{N \times M}$ with $\sigma_{i}(\mathbf{A})$ being its $i$-th \textit{singular value}. The Frobenius norm of $\mathbf{A}$ is defined as $\norm{\mathbf{A}}_{F}^{2} = \text{tr}(\mathbf{A}\mathbf{A}^{T}) = \sum_{i, j}{\mathbf{A}_{i, j}^{2}} = \sum_{i}{\sigma_{i}^{2}}$. The \textit{stable rank} \cite{Vasershtein1971} of $\mathbf{A}$ is then defined as:
\begin{equation}
    \text{sr}(\mathbf{A}) = \frac{\norm{\mathbf{A}}_{F}^{2}}{\norm{\mathbf{A}}^{2}} = \frac{\sum_{i}{\sigma_{i}^{2}}}{\max{\sigma_{i}^{2}}}
\end{equation}
and $\text{sr}(\mathbf{A}) \leq \text{rank}(\mathbf{A})$.

For $q_{\mathbf{S}} = \text{sr}(\mathbf{S})$, the loss function can be extended:
\begin{equation}
    L_{\text{sr}}(\hat{\mathbf{y}}, \mathbf{t}) = L(\hat{\mathbf{y}}, \mathbf{t}) + \alpha_{\text{sr}}\left(C - q_{\mathbf{S}}\right)
\end{equation}
where $C$ can be interpreted as the maximum possible rank of the shared embedding matrix.

\subsection{Von Neumann entropy}
It is possible to approach independence from the point of view of information theory. With this setup, \textit{von Neumann entropy} could be used to nudge the shared embedding matrix to contain independent vector components. For a \textit{density matrix} written in the basis of its eigenvectors, the von Neumann entropy is defined as:
\begin{equation}
   V(\mathbf{A}) = -\sum_{i}{\sigma_{i}^{2}\ln{\sigma_{i}^{2}}} 
\end{equation}
Let $\mathbf{R}_{D \times C}$ be defined as $\mathbf{S}$ normalized along the dimension of the shared embedding space, such that for the $i$-th row of $\mathbf{R}$ we have $\sum_{j}{\mathbf{R}_{i, j}} = 1$. In other words, $\mathbf{R}$ is the result of normalizing the rows of $\mathbf{S}$. Then, for $v_{\mathbf{R}} = V(\mathbf{R})$ we can modify the vanilla loss to make use of the von Neumann entropy:
\begin{equation}
    L_{\text{vN}}(\hat{\mathbf{y}}, \mathbf{t}) = L(\hat{\mathbf{y}}, \mathbf{t}) - \alpha_{\text{vN}}v_{\mathbf{R}}
\end{equation}
\subsection{Sparse attention}
In a procedure orthogonal to inducing structure in the shared embedding matrix, one can also restrict the way in which the actual shared embeddings are combined to form the processed embeddings. One drawback of the standard attention mechanism is that it assigns non-zero weights to all the value vectors. This means that even components with marginal similarity to the keys are present in the final linear combinations. A potential solution would be to make the output of the attention mechanism not rely on values with small similarity scores. In order to keep the whole mechanism differentiable, we adopt the $\alpha$-entmax method \cite{Peters2019}. Let us denote the $d$-probability simplex by $\triangle^{d} = \{\mathbf{p} \in \mathbb{R}^{d}: \mathbf{p} \geq 0, \norm{p}_{1} = 1\}$. Sparsemax \cite{Martins2016} is defined as:
\begin{equation}
    \text{sparsemax}(\mathbf{z}) = \argmin_{\mathbf{p} \in \triangle^{d}}{\norm{\mathbf{p} - \mathbf{z}}^{2}}
\end{equation}
A family of Tsallis $\alpha$-entropies \cite{Tsallis1988} can be defined for $\alpha \geq 1$ as:
\begin{equation}
    H_{\alpha}^{T}(\mathbf{p}) =
    \begin{cases}
        \frac{1}{\alpha(1 - \alpha)}\sum_{j}{\left(p_{j} - p_{j}^{\alpha}\right)}, & \alpha \neq 1\\
        H^{S}(\mathbf{p}), & \alpha = 1\\
    \end{cases}
\end{equation}
where $H^{S}(\mathbf{p}) = -\sum_{j}{p_{j}\ln{p_{j}}}$.

Finally, the $\alpha$-entmax, which can be understood as an interpolation between softmax and sparsemax, is defined as:
\begin{equation}
    \alpha\text{-entmax}(\mathbf{z}) = \argmax_{\mathbf{p} \in \triangle^{d}}{\mathbf{p}^{T}\mathbf{z}} + H_{\alpha}^{T}(\mathbf{p})
\end{equation}
Given this definition, $1$-entmax and $2$-entmax are identical to softmax and sparsemax, respectively. $\alpha$-entmax is differentiable, which also means that the value of the $\alpha$ parameter does not have to be supplied as a fixed hyperparameter as it can be learned together with other model parameters.

\section{Extended interpretability results}
\label{app:interpretability}

While the samples presented in the main paper are instructive of the ability of our models to produce interpretable shared embeddings and of the difference between them and a random assignment, it is important to present and analyze a larger number of samples. Additional samples for (a) the base shared embedding method, (b) the $1.05$-entmax sparse attention method and (c) random choice are presented below in Tables \ref{tab:interpretability-shared-embeddings-full}, \ref{tab:interpretability-sparse-attention-full}, \ref{tab:interpretability-random-full}, respectively. For each method, the analysis encompasses 5 trials. In each trial, a shared embedding is chosen at random without replacement. For the base shared embedding method and the $1.05$-entmax method, $5$ variables most similar to the chosen random shared embedding are presented. The similarity between the sampled shared embedding $\mathbf{s}_{p}$ and the $i$-th variable is measured as the cosine similarity $S_{C}$ between the shared embedding and the processed variable embedding $\mathbf{f}_{i}$ associated with this specific variable:

\begin{equation}
    S_{C}(\mathbf{s}_{p}, \mathbf{f}_{i}) = \frac{\mathbf{s}_{p} \cdot \mathbf{f}_{i}}{\norm{\mathbf{s}_{p}}\norm{\mathbf{f}_{i}}}
\end{equation}

For the random choice setup, a random choice without replacement of $5$ variables is shown.

\subsection{Shared embedding method}
Trials (Table \ref{tab:interpretability-shared-embeddings-full}):
\begin{enumerate}
    \item All the selected variables come from the same Subject Area (SA), Physics and Chemistry. Also, all the variables come from distinct datasets. This supports the view that the shared embedding method is able to identify the underlying abstract concepts behind the variables and does not necessarily form a very strong link between the shared embeddings and specific datasets. Qualitatively, the identified variables show a relatively consistent intuitive concept related to the measurement of physical phenomena or objects.
    \item $3$ variables come from the Biology SA, which forms the dominant category. The remaining $2$ variables come from the Physics and Chemistry SA. There are $4$ datasets represented, which shows that the shared embeddings are not strongly linked to specific datasets. Qualitatively, the chosen variables do represent an intuitive abstract concept related to biological phenomena. It can also be argued that the physical variables present in the choice describe natural phenomena, which, together with the biological variables, would form a relatively consistent grouping.
    \item There is a dominant category with $4$ variables in the form of Health and Medicine. All the selected variables come from different datasets. There is significant coherence in the grouping, which can also be seen in qualitative terms as the only physical variable in the selection can still be understood as describing elements of a real-world structure, similar to most of the biological variables. All in all, an intuitive biological concept can be identified.
    \item There is a dominant category, Health and Medicine, albeit not a majority category. There are $4$ distinct SA represented and $4$ datasets. The majority of the selected variables can still intuitively be interpreted as ones related to health or the biological functioning of organisms, but outliers, such as values from synthetically generated charts, are also present. Overall, the interpretation is made significantly harder by ambiguous data.
    \item The dominant SA, Physics and Chemistry, is represented by $2$ variables, so there is no majority SA, and also it is tied with Health and Medicine for the number of variables. All variables come from distinct datasets. Other than in other trials, there is a more clear split of meaning between two concepts: physical and health-related ones. There is still some intuitive overlap but the internal consistency of the variables is weaker than for the other trials.
\end{enumerate}

Overall, the shared embedding method results in similar variables which have a majority SA in $3/5$ trials, a dominant category without ties in $4/5$ trials and a dominant category with possible draws in all $5/5$ trials. Also, there is at most one repeated dataset in any of the trials. If we were to adopt a view that different versions of the same dataset effectively count as one dataset, then we would only have one trial (2) with two repeated datasets. Qualitatively, all the the selected trials display the potential of the method to identify abstract concepts from varied areas.

\begin{table}[h]
\caption{Most similar variables from the UCI-121 dataset for a random choice of a shared embedding. Shared embedding method. Variables sorted in descending order of similarity. Missing values (-) denote ambiguous data. (*) denotes inferred Subject Areas. The \textit{Remarks} column lists the most dominant Subject Area (SA), the number of SAs present and the number of distinct datasets represented.}
\label{tab:interpretability-shared-embeddings-full}
\vskip 0.15in
\begin{center}
\begin{scriptsize}
\begin{sc}
\begin{tabular}{p{0.02\linewidth} p{0.16\linewidth} p{0.3\linewidth} p{0.2\linewidth} p{0.09\linewidth}}
\toprule
No. & Dataset & Variable meaning & Subject Area & Remarks \\
\midrule
\multirow{5}{*}{(1)}
    & \multicolumn{1}{l}{MUSK (V2)} & \multicolumn{1}{l}{\textit{A distance feature of a molecule along a ray.}} & \multicolumn{1}{l}{Physics and Chemistry} & \\
    & \multicolumn{1}{l}{C. Bench (S,MvsR)} & \multicolumn{1}{l}{\textit{Energy within a frequency band, integrated over time.}} & \multicolumn{1}{l}{Physics and Chemistry} & Dom.: 5/5 \\
    & \multicolumn{1}{l}{Waveform (V1)} & \multicolumn{1}{l}{\textit{Waveform feature; contains noise
but is not all noise.}} & \multicolumn{1}{l}{Physics and Chemistry} & SAs: 1 \\
    & \multicolumn{1}{l}{MiniBooNE} & \multicolumn{1}{l}{\textit{A particle ID variable (real) for an event.}} & \multicolumn{1}{l}{Physics and Chemistry} & D-sets: 5 \\
    & \multicolumn{1}{l}{Annealing} & \multicolumn{1}{l}{-} & \multicolumn{1}{l}{Physics and Chemistry} & \\
\midrule
\multirow{5}{*}{(2)}
    & \multicolumn{1}{l}{-} & \multicolumn{1}{l}{-} & \multicolumn{1}{l}{Biology(*)} & \\
    & \multicolumn{1}{l}{MUSK (V2)} & \multicolumn{1}{l}{\textit{A distance feature of a molecule along a ray.}} & \multicolumn{1}{l}{Physics and Chemistry} & Dom.: 3/5 \\
    & \multicolumn{1}{l}{Leaves (Shape)} & \multicolumn{1}{l}{\textit{A specific feature relating to the shape of the leaf.}} & \multicolumn{1}{l}{Biology} & SAs: 2 \\
    & \multicolumn{1}{l}{MUSK (V1)} & \multicolumn{1}{l}{\textit{A distance feature of a molecule along a ray.}} & \multicolumn{1}{l}{Physics and Chemistry} & D-sets: 4 \\
    & \multicolumn{1}{l}{Leaves (Shape)} & \multicolumn{1}{l}{\textit{A specific feature relating to the shape of the leaf.}} & \multicolumn{1}{l}{Biology} \\
\midrule
\multirow{5}{*}{(3)}
    & \multicolumn{1}{l}{Dermatology} & \multicolumn{1}{l}{\textit{Thinning of the suprapapillary epidermis.}} & \multicolumn{1}{l}{Health and Medicine} & \\
    & \multicolumn{1}{l}{SPECT Heart} & \multicolumn{1}{l}{\textit{Binary feature of cardiac CT images.}} & \multicolumn{1}{l}{Health and Medicine} & Dom.: 4/5 \\
    & \multicolumn{1}{l}{MiniBooNE} & \multicolumn{1}{l}{\textit{A particle ID variable (real) for an event}} & \multicolumn{1}{l}{Physics and Chemistry} & SAs: 2 \\
    & \multicolumn{1}{l}{Haberman} & \multicolumn{1}{l}{\textit{Number of positive axillary nodes detected.}} & \multicolumn{1}{l}{Health and Medicine} & D-sets: 5 \\
    & \multicolumn{1}{l}{Heart Dis. (CH)} & \multicolumn{1}{l}{-} & \multicolumn{1}{l}{Health and Medicine} \\
\midrule
\multirow{5}{*}{(4)}
    & \multicolumn{1}{l}{Arrhythmia} & \multicolumn{1}{l}{-} & \multicolumn{1}{l}{Health and Medicine} & \\
    & \multicolumn{1}{l}{-} & \multicolumn{1}{l}{-} & \multicolumn{1}{l}{Biology(*)} & Dom.: 2/5 \\
    & \multicolumn{1}{l}{Arrhythmia} & \multicolumn{1}{l}{-} & \multicolumn{1}{l}{Health and Medicine} & SAs: 4 \\
    & \multicolumn{1}{l}{Synth. Control} & \multicolumn{1}{l}{\textit{Point value on synthetically generated control chart.}} & \multicolumn{1}{l}{Other} & D-sets: 4 \\
    & \multicolumn{1}{l}{MiniBooNE} & \multicolumn{1}{l}{\textit{A particle ID variable (real) for an event}} & \multicolumn{1}{l}{Physics and Chemistry} & \\
\midrule
\multirow{5}{*}{(5)}
    & \multicolumn{1}{l}{Annealing} & \multicolumn{1}{l}{-} & \multicolumn{1}{l}{Physics and Chemistry} & \\
    & \multicolumn{1}{l}{Breast Cancer} & \multicolumn{1}{l}{\textit{Whether irradiation was used.}} & \multicolumn{1}{l}{Health and Medicine} & Dom.: 2/5 \\
    & \multicolumn{1}{l}{OR of H. Digits} & \multicolumn{1}{l}{\textit{Preprocessed feature of a digit image.}} & \multicolumn{1}{l}{Computer Science} & SAs: 3 \\
    & \multicolumn{1}{l}{MUSK (V2)} & \multicolumn{1}{l}{\textit{A distance feature of a molecule along a ray.}} & \multicolumn{1}{l}{Physics and Chemistry} & D-sets: 5 \\
    & \multicolumn{1}{l}{Primary Tumor} & \multicolumn{1}{l}{\textit{Whether sample is related to supraclavicular LNs.}} & \multicolumn{1}{l}{Health and Medicine} \\
\bottomrule
\end{tabular}
\end{sc}
\end{scriptsize}
\end{center}
\vskip -0.1in
\end{table}

\vfill
\eject

\subsection{$1.05$-entmax method}
Trials (Table \ref{tab:interpretability-sparse-attention-full}):
\begin{enumerate}
    \item $3$ variables come from the Health and Medicine SA and form a dominant category. All the variables are from distinct datasets. Quantitatively, the method shows potential to identify notions related to health. The qualitative analysis is hindered by the ambiguity of the data, however, one might still identify an intuitive concept relating to diseases or a broader one relating to living organisms.
    \item A failure case: all the variables are from different SA and so, there is no reliable dominant category. All the variables come from distinct datasets. For this specific trial, no underlying concept can be easily identified. 
    \item There is a dominant category, Physics and Chemistry, with $2$ variables, but it is tied for the lead with another SA, Health and Medicine, in terms of the number of identified variables. All the selected variables come from distinct datasets. There are two underlying intuitive concepts: a physical one and one related to health.
    \item We do have a dominant SA, Biology, with $2$ variables, but again, there is another category with the same number of identified variables - Health and Medicine. Also, we see that for this trial, there are 2 repeated datasets. The intuitive meaning behind the variables from this trial can be interpreted as describing living things but more details are occluded by the fact that all the variables for the Health and Medicine SA are ambiguous.
    \item Biology is the dominant category with $2$ representatives, but the Health and Medicine SA has the same number of identified variables. All the variables come from distinct datasets. Qualitatively, the underlying concept can be identified as a description of a real-world structure or a point on a larger representation of a phenomenon. With this interpretation, even the variable coming from the Other SA fits the concept. 
\end{enumerate}

The $1.05$-etnmax sparse attention method identifies variables in a distinctly different way than the base shared embedding method. Namely, there are far less cases with majority SAs and far more outcomes where the dominant category is tied for the lead with another SA as far as the number of identified variables is concerned. The sparse attention method still prefers variables from distinct datasets and does not seem to very strongly link a particular shared embedding to a concrete dataset. At the same time, both the quantitative metrics and the qualitative assessment suggest that it is the base shared embedding method that more successfully delineates between abstract concepts. 

\begin{table}[h]
\caption{Most similar variables from the UCI-121 dataset for a random choice of a shared embedding. $1.05$-entmax sparse attention method with embeddings initialized from $\mathcal{N}(0, 1)$. Variables sorted in descending order of similarity. Missing values (-) denote ambiguous data. (*) denotes inferred Subject Areas. The \textit{Remarks} column lists the most dominant Subject Area (SA), the number of SAs present and the number of distinct datasets represented.}
\label{tab:interpretability-sparse-attention-full}
\vskip 0.15in
\begin{center}
\begin{scriptsize}
\begin{sc}
\begin{tabular}{p{0.02\linewidth} p{0.16\linewidth} p{0.3\linewidth} p{0.2\linewidth} p{0.09\linewidth}}
\toprule
No. & Dataset & Variable meaning & Subject Area & Remarks \\
\midrule
\multirow{5}{*}{(1)}
    & \multicolumn{1}{l}{Br. Cancer WI (D.)} & \multicolumn{1}{l}{\textit{Mean compactness of the cell nuclei in the image.}} & \multicolumn{1}{l}{Health and Medicine} & \\
    & \multicolumn{1}{l}{Thyroid Disease} & \multicolumn{1}{l}{-} & \multicolumn{1}{l}{Health and Medicine} & Dom.: 3/5 \\
    & \multicolumn{1}{l}{Arrhythmia} & \multicolumn{1}{l}{-} & \multicolumn{1}{l}{Health and Medicine} & SAs: 3 \\
    & \multicolumn{1}{l}{Leaves (Shape)} & \multicolumn{1}{l}{\textit{A specific feature relating to the shape of the leaf.}} & \multicolumn{1}{l}{Biology} & D-sets: 5 \\
    & \multicolumn{1}{l}{Synth. Control} & \multicolumn{1}{l}{\textit{Point value on synthetically generated control chart.}} & \multicolumn{1}{l}{Other} & \\
\midrule
\multirow{5}{*}{(2)}
    & \multicolumn{1}{l}{-} & \multicolumn{1}{l}{\textit{-}} & \multicolumn{1}{l}{Biology(*)} & \\
    & \multicolumn{1}{l}{Connect-4} & \multicolumn{1}{l}{\textit{Which of the players has taken position d5.}} & \multicolumn{1}{l}{Games} & Dom.: 1/5 \\
    & \multicolumn{1}{l}{Synth. Control} & \multicolumn{1}{l}{\textit{Point value on synthetically generated control chart.}} & \multicolumn{1}{l}{Other} & SAs: 5 \\
    & \multicolumn{1}{l}{Arrhythmia} & \multicolumn{1}{l}{-} & \multicolumn{1}{l}{Health and Medicine} & D-sets: 5 \\
    & \multicolumn{1}{l}{MUSK (V2)} & \multicolumn{1}{l}{\textit{A distance feature of a molecule along a ray.}} & \multicolumn{1}{l}{Physics and Chemistry} & \\
\midrule
\multirow{5}{*}{(3)}
    & \multicolumn{1}{l}{MUSK (V1)} & \multicolumn{1}{l}{\textit{A distance feature of a molecule along a ray.}} & \multicolumn{1}{l}{Physics and Chemistry} & \\
    & \multicolumn{1}{l}{Wine} & \multicolumn{1}{l}{\textit{Flavanoids.}} & \multicolumn{1}{l}{Physics and Chemistry} & Dom.: 2/5 \\
    & \multicolumn{1}{l}{Statlog (V. Silh.)} & \multicolumn{1}{l}{\textit{Elongatedness of a silhouette of a vehicle.}} & \multicolumn{1}{l}{Other} & SAs: 3 \\
    & \multicolumn{1}{l}{Br. Cancer WI (P.)} & \multicolumn{1}{l}{\textit{Mean texture of the cell nuclei in the image.}} & \multicolumn{1}{l}{Health and Medicine} & D-sets: 5 \\
    & \multicolumn{1}{l}{Arrhythmia} & \multicolumn{1}{l}{-} & \multicolumn{1}{l}{Health and Medicine} & \\
\midrule
\multirow{5}{*}{(4)}
    & \multicolumn{1}{l}{Leaves (Shape)} & \multicolumn{1}{l}{\textit{A specific feature relating to the shape of the leaf.}} & \multicolumn{1}{l}{Biology} & \\
    & \multicolumn{1}{l}{Arrhythmia} & \multicolumn{1}{l}{-} & \multicolumn{1}{l}{Health and Medicine} & Dom.: 2/5 \\
    & \multicolumn{1}{l}{Arrhythmia} & \multicolumn{1}{l}{-} & \multicolumn{1}{l}{Health and Medicine} & SAs: 3 \\
    & \multicolumn{1}{l}{Leaves (Shape)} & \multicolumn{1}{l}{\textit{A specific feature relating to the shape of the leaf.}} & \multicolumn{1}{l}{Biology} & D-sets: 3 \\
    & \multicolumn{1}{l}{C. Bench (S,MvsR)} & \multicolumn{1}{l}{\textit{Energy within a frequency band, integrated over time.}} & \multicolumn{1}{l}{Physics and Chemistry} & \\
\midrule
\multirow{5}{*}{(5)}
    & \multicolumn{1}{l}{Horse Colic} & \multicolumn{1}{l}{\textit{Temperature of extremities.}} & \multicolumn{1}{l}{Biology} & \\
    & \multicolumn{1}{l}{Mol. Biol. (PGS)} & \multicolumn{1}{l}{\textit{Position -50 in the DNA sequence.}} & \multicolumn{1}{l}{Biology} & Dom.: 2/5 \\
    & \multicolumn{1}{l}{Lung Cancer} & \multicolumn{1}{l}{-} & \multicolumn{1}{l}{Health and Medicine} & SAs: 3 \\
    & \multicolumn{1}{l}{Synth. Control} & \multicolumn{1}{l}{\textit{Point value on synthetically generated control chart.}} & \multicolumn{1}{l}{Other} & D-sets: 5 \\
    & \multicolumn{1}{l}{Heart Dis. (VALB)} & \multicolumn{1}{l}{\textit{Maximum heart rate achieved.}} & \multicolumn{1}{l}{Health and Medicine} & \\
\bottomrule
\end{tabular}
\end{sc}
\end{scriptsize}
\end{center}
\vskip -0.1in
\end{table}

\begin{table}[h!]
\caption{Random choice of variables from the UCI-121 dataset. Missing values (-) denote ambiguous data. (*) denotes inferred Subject Areas. The \textit{Remarks} column lists the most dominant Subject Area (SA), the number of SAs present and the number of distinct datasets represented.}
\label{tab:interpretability-random-full}
\vskip 0.15in
\begin{center}
\begin{scriptsize}
\begin{sc}
\begin{tabular}{p{0.02\linewidth} p{0.16\linewidth} p{0.3\linewidth} p{0.2\linewidth} p{0.09\linewidth}}
\toprule
No. & Dataset & Variable meaning & Subject Area & Remarks \\
\midrule
\multirow{5}{*}{(1)}
    & \multicolumn{1}{l}{Audiology (S.)} & \multicolumn{1}{l}{-} & \multicolumn{1}{l}{Health and Medicine} & \\
    & \multicolumn{1}{l}{-} & \multicolumn{1}{l}{-} & \multicolumn{1}{l}{Biology(*)} & Dom.: 2/5 \\
    & \multicolumn{1}{l}{Synth. Control} & \multicolumn{1}{l}{\textit{Point value on synthetically generated control chart.}} & \multicolumn{1}{l}{Other} & SAs: 4 \\
    & \multicolumn{1}{l}{Tic-Tac-Toe End.} & \multicolumn{1}{l}{\textit{State of the bottom-left square at the end of a game.}} & \multicolumn{1}{l}{Games} & D-sets: 5 \\
    & \multicolumn{1}{l}{Heart Dis. (CH)} & \multicolumn{1}{l}{-} & \multicolumn{1}{l}{Health and Medicine} & \\
\midrule
\multirow{5}{*}{(2)}
    & \multicolumn{1}{l}{C. Bench (S,MvsR)} & \multicolumn{1}{l}{\textit{Energy within a frequency band, integrated over time.}} & \multicolumn{1}{l}{Physics and Chemistry} & \\
    & \multicolumn{1}{l}{MUSK (V2)} & \multicolumn{1}{l}{\textit{A distance feature of a molecule along a ray.}} & \multicolumn{1}{l}{Physics and Chemistry} & Dom.: 2/5 \\
    & \multicolumn{1}{l}{Statlog (Image S.)} & \multicolumn{1}{l}{-} & \multicolumn{1}{l}{Other} & SAs: 4 \\
    & \multicolumn{1}{l}{Yeast} & \multicolumn{1}{l}{\textit{Score of discriminant analysis of proteins.}} & \multicolumn{1}{l}{Biology} & D-sets: 5 \\
    & \multicolumn{1}{l}{Arrhythmia} & \multicolumn{1}{l}{-} & \multicolumn{1}{l}{Health and Medicine} & \\
\midrule
\multirow{5}{*}{(3)}
    & \multicolumn{1}{l}{Mol. Biol. (SGS)} & \multicolumn{1}{l}{\textit{Position +23 in the DNA sequence.}} & \multicolumn{1}{l}{Biology} & \\
    & \multicolumn{1}{l}{MUSK (V2)} & \multicolumn{1}{l}{\textit{A distance feature of a molecule along a ray.}} & \multicolumn{1}{l}{Physics and Chemistry} & Dom.: 2/5 \\
    & \multicolumn{1}{l}{Ozone Level} & \multicolumn{1}{l}{\textit{Precipitation.}} & \multicolumn{1}{l}{Climate and Env.} & SAs: 3 \\
    & \multicolumn{1}{l}{Soybean (Large)} & \multicolumn{1}{l}{\textit{Type of seed treatment (e.g. fungicide).}} & \multicolumn{1}{l}{Biology} & D-sets: 5 \\
    & \multicolumn{1}{l}{LR Spectrometer} & \multicolumn{1}{l}{\textit{Specific flux measurement for the red band.}} & \multicolumn{1}{l}{Physics and Chemistry} & \\
\midrule
\multirow{5}{*}{(4)}
    & \multicolumn{1}{l}{Libras Movement} & \multicolumn{1}{l}{\textit{Coordinate abcissa of the 19th point.}} & \multicolumn{1}{l}{Other} & \\
    & \multicolumn{1}{l}{Arrhythmia} & \multicolumn{1}{l}{-} & \multicolumn{1}{l}{Health and Medicine} & Dom.: 3/5 \\
    & \multicolumn{1}{l}{Pittsburgh Bridges} & \multicolumn{1}{l}{\textit{Purpose of the bridge.}} & \multicolumn{1}{l}{Other} & SAs: 2 \\
    & \multicolumn{1}{l}{Trains} & \multicolumn{1}{l}{-} & \multicolumn{1}{l}{Other} & D-sets: 5 \\
    & \multicolumn{1}{l}{Dermatology} & \multicolumn{1}{l}{\textit{Clinical attributes: definite borders.}} & \multicolumn{1}{l}{Health and Medicine} & \\
\midrule
\multirow{5}{*}{(5)}
    & \multicolumn{1}{l}{MUSK (V1)} & \multicolumn{1}{l}{\textit{A distance feature of a molecule along a ray.}} & \multicolumn{1}{l}{Physics and Chemistry} & \\
    & \multicolumn{1}{l}{-} & \multicolumn{1}{l}{-} & \multicolumn{1}{l}{Biology(*)} & Dom.: 2/5 \\
    & \multicolumn{1}{l}{Haberman} & \multicolumn{1}{l}{\textit{Number of positive axillary nodes detected.}} & \multicolumn{1}{l}{Health and Medicine} & SAs: 3 \\
    & \multicolumn{1}{l}{MUSK (V2)} & \multicolumn{1}{l}{\textit{A distance feature of a molecule along a ray.}} & \multicolumn{1}{l}{Physics and Chemistry} & D-sets: 5 \\
    & \multicolumn{1}{l}{Mol. Biol. (PGS)} & \multicolumn{1}{l}{\textit{Position -22 in the DNA sequence.}} & \multicolumn{1}{l}{Biology} & \\
\bottomrule
\end{tabular}
\end{sc}
\end{scriptsize}
\end{center}
\vskip -0.1in
\end{table}

\subsection{Random choice}
\label{app:interpretability-random}
In order to account for the statistical properties of the UCI-121 dataset, we perform an analysis where the selected variables are actually randomly sampled without repetition from the dataset. If the dataset is not heavily skewed toward the concepts identified by either of our methods, it is natural to assume that we will see a lot more variability in the selection. For a random choice of variables, one could expect not to see majority SAs, or at least see them infrequently. Similarly, the expectation would be to see more SAs within each trial than is the case for our methods. Also, a random assignment would result in very frequent situations where all the variables come from distinct datasets. Conversely, for the base shared embedding method and the $1.05$-entmax method the expectation would be that the variables most similar to a given shared embedding would be more likely to come from the same dataset. Table \ref{tab:interpretability-random-full} summarizes the results for the random choice of variables. Indeed, there is only one trail with a majority category, but on inspection the identified SA is Other, which is a blanket category for a range of datasets representing different concepts. Apart from this special case, there are no other majority categories in trials. This suggests significantly weaker interpretability than for the base shared embedding method. The sparse attention method does show similar levels of dominant categories, however, with a crucial distinction. In the sparse attention approach, all the non-majority cases bar one had a tie for the dominant category, suggesting that the method was able to identify concepts better than random choice, with the assignment to two competing concepts. Overall, for the sparse attention method, $4/5$ trials either had a majority category or a tied dominant category. For random choice, excluding the kitchen sink Other SA, a majority category or a draw between two competing categories occurs in $2/5$ trials. This suggests that for random choice there is less concentration in SAs. Also, there are visibly more SAs represented than for the shared embedding method. The shared embedding method has an average of $2.4$ SAs per trial. The same metric for random choice stands at $3.2$. Qualitatively, random choice does result in an assortment of more than two distinct concepts for a given trail rather than in the identification of an abstract notion or two such notions, which is a frequent situation for the shared embedding method and $1.05$-entmax methods.


\section{Architecture of the proposed method}
\label{app:architecture}

\begin{figure}[H]
\vskip 0.2in
\begin{center}
\centerline{\includegraphics[width=\columnwidth]{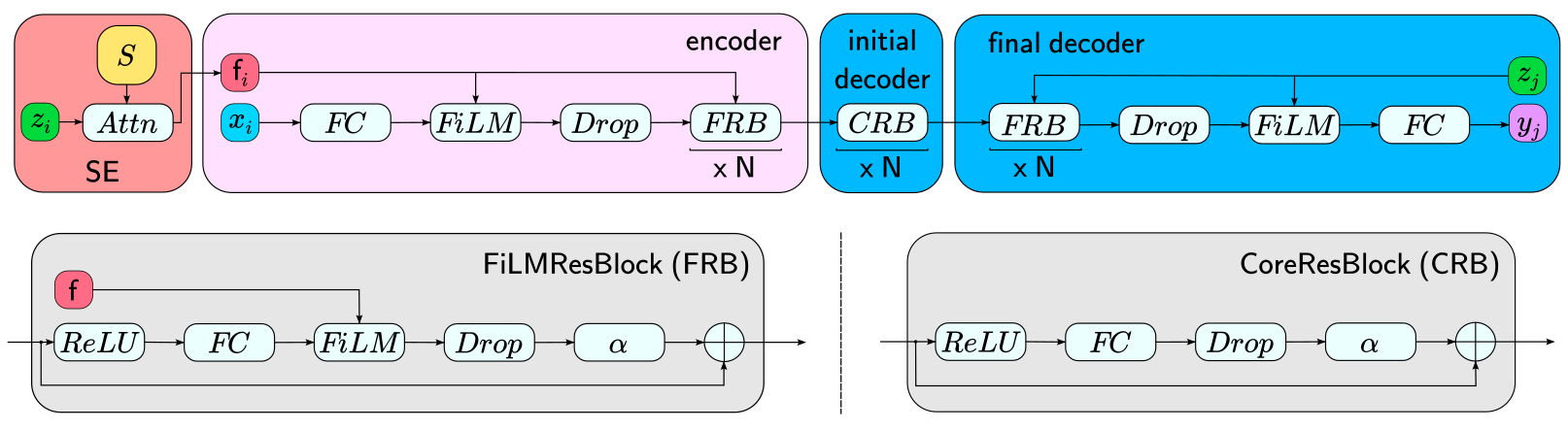}}
\caption{Architecture of the \textit{shared variable embeddings} method. SE - shared embeddings, Attn - attention, S - shared embedding matrix, FC - fully connected layers, FiLM - layers proposed by \cite{Perez2018}, Drop - dropout, ReLU - rectified linear units.}
\label{fig:architecture}
\end{center}
\vskip -0.2in
\end{figure}

\section{Extended ablations}
\label{app:hyperparameter}

Our hyperparameter choice, performed largely before the training of SVE, follows one overriding goal: to match the hyperparameters used by the TOM baseline \cite{Meyerson2021} wherever possible. For all the hyperparameter choices, apart from one, we strictly adhere to the values for the TOM baseline. For instance, the dimensionality of our initial variable embedding space, the dimensionality of the internal representations of the encoders/decoder, the number of encoders/decoder layers, the dropout rate, the learning rate, the weight decay, the stop criteria and other hyperparameters are all the same as for TOM.

Our architecture with spare attention introduces three hyperparameters not present in the TOM baseline: the $\alpha$-entmax parameter, the number of shared variable embeddings and the dimensionality of the shared embedding space. We let SGD optimize $\alpha$, and choose its starting value according to the ablation presented in the paper. The dimensionality of the shared embedding space needs to match the dimensionality of the raw variable embeddings, which restricts it to a value for the latter from the TOM baseline. The number of the shared embeddings is chosen equal to their dimensionality in order for the shared embedding matrix to be square, which is required for the analysis of the \textit{orthogonality} restriction.

The only hyperparameter present in TOM for which we choose a different value is the standard deviation of the distribution from which the variable embeddings are initialized. In TOM this distribution is $\mathcal{N}(0, 10^{-3})$. In SVE, it is $\mathcal{N}(0, 1)$. This change is dictated by the fact that SVE introduces an attention mechanism, which relies on the computation of dot products between raw and shared embeddings. Keeping the standard deviation as in TOM results in vanishing dot product values and constant output from the softmax in the attention mechanism. In order to circumvent that, we choose the standard deviation based on the ablation reported in the main paper.

The goal of our choice of hyperparameters is to explicitly follow the training protocol of the TOM baseline as closely as possible without performing an extensive search of suitable hyperparemeter values. Specifically, we do not use cherry-picked hyperparameters to achieve the levels of accuracy and training time reported in Section \ref{sec:experiments}. It is nevertheless of interest to verify the sensitivity of the obtained results to the hyperparameters. Towards this end, we have performed a more extensive search to determine how SVE behaves for different hyperparameter levels. We have found that there are hyperparameter combinations for which our method performs materially better in terms of classification accuracy and for which it also requires significantly less training steps than reported in the paper.

Since an exhaustive grid search would have been prohibitive, we decided on using our $1.05$-entmax sparse attention model as a starting point and have analyzed the sensitivity of the results to hyperparameter manipulations. Concretely, we check the impact of changing \textit{each one} of the hyperparameters, other than those that are already discussed in the main body of the paper.

We start with the analysis of how both the dimensionality of the raw embedding space ($C$) and that of the shared embedding space ($D$) influence test set accuracy. The results are presented in Table \ref{tab:emb-dim}.

\begin{table}[h]
\caption{Test set accuracy for specific combinations of the dimensionality of the raw embedding space ($C$) and the shared embedding space ($D$).}
\label{tab:emb-dim}
\centering
\begin{tabular}{|c| c c c c c c|} 
 \hline
 $C \backslash D$ & 32 & 64 & 128 & 256 & 512 & 1\,024 \\ [0.5ex] 
 \hline
 32 & 79.5 & 80.4 & 79.8 & 81.1 & 79.1 & 80.8 \\ 
 64 & 79.1 & 80.7 & 80.5 & 80.2 & 80.2 & 80.2 \\
 128 & 78.1 & 76.9 & 81.9 & 79.4 & 79.8 & 78.5 \\
 256 & 76.0 & 77.1 & 79.3 & 80.5 & 80.5 & 79.7 \\
 512 & 72.9 & 77.3 & 78.7 & 79.8 & 77.4 & 77.1 \\
 1\,024 & 73.7 & 73.6 & 75.6 & 73.3 & 74.5 & 76.3 \\ [1ex]
 \hline
\end{tabular}
\end{table}

Overall, we see that the test set accuracy is relatively insensitive to the choice of the dimensionality of the shared embedding space. The choice of the raw embedding space seems more pertinent to the performance of the model and we see a drop-off in performance for $C \geq 512$.

A subtle question is whether using one dimensionality for input and target embeddings, which is required in the SVE approach, is valid. In our analysis of the dimensionality of the raw and shared embedding dimensions, we find that for reasonable choices of $C$ and $D$ the performance does not suffer. The intuition behind this is that even for the dimension of $32$, a real-values vector with $32$ components is able to encode both up to $262$ input features and $100$ classes for the UCI-121 dataset. If anything, the dimensionality may be a bit of an overkill for the output, which most frequently has far less than $200$ classes. In a situation where the imbalance between the inputs and targets would be extreme, the common dimensionality could conceivably be a problem but this is not something we observe in practice.

Similarly to the dimensionality of the raw and shared embedding spaces, we consider the impact of the dimensionality of the \textit{latent space} ($H$), i.e. the dimensionality of the internal representations of the encoders and the decoder. This is shown in Table~\ref{tab:latent-dim}.

\begin{table}[h]
\caption{Test set accuracy for specific dimensionality of the latent space ($H$).}
\label{tab:latent-dim}
\centering
\begin{tabular}{||c c c c c||} 
 \hline
 $H$ & 32 & 64 & 128 & 256 \\ [0.5ex] 
 \hline\hline
 mean test set acc & 80.6 & 79.4 & 81.9 & 80.0 \\ [1ex]
 \hline
\end{tabular}
\end{table}

We stop our analysis at $256$ as this is the highest dimension that fits in the memory of the machine we use for training. Again, the results are relatively insensitive to the choice of $H$, with the best results for moderate levels consistent with those used in the paper.

We next focus on the number of layers of our encoders and the decoder. We assume an equal number of layers for the three networks (two encoders and the decoder). The results are presented in Table \ref{tab:layers}.

\begin{table}[h]
\caption{Test set accuracy for different numbers of network layers.}
\label{tab:layers}
\centering
\begin{tabular}{||c c c c c||} 
 \hline
 layers & 5 & 10 & 15 & 20 \\ [0.5ex] 
 \hline\hline
 mean test set acc & 80.1 & 81.9 & 79.7 & 80.0 \\ [1ex]
 \hline
\end{tabular}
\end{table}

We see limited sensitivity to the choice of the number of layers. In particular, it does not seem that increasing the number of layers beyond the value used in the paper (10) is beneficial in terms of performance.

Further on, let us consider the influence of dropout on the results from SVE - Table \ref{tab:dropout}.
\begin{table}[h]
\caption{Test set accuracy for different levels of dropout.}
\label{tab:dropout}
\centering
\begin{tabular}{||c c c c c c c||} 
 \hline
 dropout & 0.0 & 0.1 & 0.2 & 0.3 & 0.4 & 0.5 \\ [0.5ex] 
 \hline\hline
 mean test set acc & 81.9 & 81.5 & 82.0 & 81.9 & \textbf{82.2} & 80.8 \\ [1ex]
 \hline
\end{tabular}
\end{table}
Notably, we see that we are able to obtain results better than those presented in the main body of the paper. By carefully tuning the dropout level, it is possible to visibly outperform the model reported in the paper. Additionally, the introduction of dropout considerably lowers the number of training steps required. For instance, for $0.4$ dropout the reported accuracy is achieved after only $8\,700$ steps, relative to $10\,300$ steps from the paper, which is a $15.5\%$ decline.

The choice of the learning rate turns out to be one of the few factors which strongly drive the performance of SVE - Table~\ref{tab:learning-rate}.

\begin{table}[h]
\caption{Test set accuracy for different learning rate levels.}
\label{tab:learning-rate}
\centering
\begin{tabular}{||c c c c c c||} 
 \hline
 learning rate & $10^{-1}$ & $10^{-2}$ & $10^{-3}$ & $10^{-4}$ & $10^{-5}$ \\ [0.5ex] 
 \hline\hline
 mean test set acc & 42.2 & 79.0 & 81.9 & 80.7 & 76.2 \\ [1ex]
 \hline
\end{tabular}
\end{table}

Having said that, for choices outside of the extremes, the performance of SVE is still relatively stable.

A similar remark holds for the weight decay parameter, which is displayed in Table \ref{tab:weight-decay}.

\begin{table}[h]
\caption{Test set accuracy for different weight decay levels.}
\label{tab:weight-decay}
\centering
\begin{tabular}{||c c c c||} 
 \hline
 weight decay & $10^{-4}$ & $10^{-5}$ & $10^{-6}$ \\ [0.5ex] 
 \hline\hline
 mean test set acc & 73.7 & 81.9 & 79.0 \\ [1ex]
 \hline
\end{tabular}
\end{table}

\section{UCI-121 dataset}
\label{app:uci-121}
The UCI-121 dataset is a collection of 121 classification datasets from the UCI Machine Learning Repository. This specific collection was first introduced in \cite{Delgado2014}, based on the UCI repository itself \cite{Kelly2023}. The relative unfamiliarity of the UCI-121 datasets stems from the fact that it comprises \textit{disjoint}, seemingly unrelated tasks and as such has so far not been extensively explored in the MTL literature. As far as the overall dataset itself is concerned, each of the 121 tasks (constituent datasets) has its own number of input features (variables), ranging from 3 to 262, and its own number of classes, ranging from 2 to 100. The names of the constituent datasets are given in the file with per-task test set results: \url{https://github.com/anonomous678876/anonymous/blob/main/results-per-dataset.xlsx}. The overall number of individual input variables in the whole dataset is $3\,490$, which precludes an exhaustive description of them in the paper. Examples of datasets and variables are given in Section \ref{sec:interpretability} and in Appendix \ref{app:interpretability}.

\section{Performance on concrete UCI-121 tasks}
\label{app:tasks}

In order to provide a more fine-grained assessment of the performance of the proposed method, we have recorded the per task accuracy both for the baseline and for SVE in the sparse attention version. We provide these per task accuracy levels for the UCI-121 dataset in the following file: \url{https://github.com/anonomous678876/anonymous/blob/main/results-per-dataset.xlsx}. In general, SVE does not necessarily perform similarly to TOM on the same tasks, and the differences in accuracies can be significant either way. SVE specializes on its own set of tasks, more than making up for the tasks where it underperforms the baseline.

\section{Variable embeddings commonly shared by tasks}
\label{app:sves-shared-commonly}

A discussion of SVEs commonly shared by tasks is difficult in the absence of concrete definitions of what \textit{commonly} and \textit{shared} mean. We have performed an additional investigation into this matter. In this investigation, we assume that a shared variable embedding is shared across tasks if for each of these tasks at least one of the task variables gets an attention probability score - the attention score after softmax - of $> 0.1$. This means that we focus on $9.3\%$ out of all the possible $121 \times 128$ task/shared embedding pairings. We further assume that the sharing is common if the shared embedding is among the top five most shared embeddings.

In more concrete terms, the procedure looks as follows. From all the $121 \times 128$ task/shared embedding attention probability scores we only count those $> 0.1$. The counting is done per shared variable embedding, which results in $128$ task counts where each tasks count represents the number of tasks for which at least one variable satisfies the attention probability score condition relative to the given shared embedding. From this list of $128$ counts we choose the $5$ largest ones. This results in a list of $5$ shared embeddings, along with their task counts. For each out of those $5$ shared embeddings, we find the top $5$ tasks with highest maximum similarity scores with this shared embedding. This gives us the final result: a list of $5$ most commonly shared variable embeddings along with the tasks that share them the most.

We present these most commonly shared variable embeddings for our vanilla SVE architecture, since this is the architecture for which the interpretability is the strongest. Each shared variable embedding is represented by a list of tasks which share them the most. These results are rendered in Table \ref{tab:sves-commonly-shared}.

\begin{table}[h]
\caption{SVEs commonly shared by tasks.}
\label{tab:sves-commonly-shared}
\begin{center}
\begin{tabular}{||c c c c||} 
 \hline
 SVE & 52 & 71 & 102\\ [0.5ex] 
 \hline\hline
 task 1 & image-segmentation & statlog-heart & hill-valley \\ [1ex]
 task 2 & libras & libras & conn-bench-vowel-deterding \\ [1ex]
 task 3 & low-res-spect & plant-shape & oocytes-merluccius-states-2f \\ [1ex]
 task 4 & musk-2 & musk-1 & horse-colic \\ [1ex]
 task 5 & optical & car & musk-2 \\ [1ex]
 \hline
\end{tabular}
\end{center}
\vspace{0.1cm}
\begin{center}
\begin{tabular}{||c c c||} 
 \hline
 SVE & 97 & 35 \\ [0.5ex] 
 \hline\hline
 task 1 & arrhythmia & statlog-australian-credit \\ [1ex]
 task 2 & low-res-spect & hill-valley \\ [1ex]
 task 3 & statlog-german-credit & monks-3 \\ [1ex]
 task 4 & ringnorm & chess-krvkp \\ [1ex]
 task 5 & musk-2 & musk-2 \\ [1ex]
 \hline
\end{tabular}
\end{center}
\end{table}

We are able to determine that these most commonly shared variable embeddings are shared across a variety of tasks from different SAs. This seems to hold for various levels of the attention probability threshold. We have checked levels between $0.05$ and $0.5$ and have found these results to largely hold unchanged.

\section{Additional datasets}
\label{app:additional-datasets}

Our evaluation on the UCI-121 dataset is dictated by two factors: 1) that we want to make a direct comparison with the TOM baseline and this baseline was only ever trained on one real-world dataset, UCI-121, 2) that there is a lack of high-quality classification datasets related to MTL on \textit{disjoint} tasks. 

Having said that, in order to further support the generality of our results, we have performed additional experiments on another classification dataset that fits our needs - the classification part of the Penn Machine Learning Benchmarks (PMLB) dataset: \url{https://epistasislab.github.io/pmlb/index.html}. This dataset provides $164$ classification tasks. From those, we filter out datasets with either very high numbers of features ($\geq 1\,000$) or very large numbers of examples ($\geq 500\,000$). This leaves us with $159$ classification datasets. It has to be noted that these datasets have some overlap with UCI-121. By manual inspection, we were able to determine that out of the $159$ selected datasets $82$ are not present in UCI-121. Still, the number of new tasks is significant enough to provide a meaningful new comparison.

We train both SVE in the $1.05$-entmax version and the TOM baseline and evaluate them on PMLB (experiment repeated twice). The results are presented in Table \ref{tab:pmlb-classification}.

\begin{table}[h]
\caption{Test set accuracy on the PMLB Classification dataset (experiment repeated twice).}
\label{tab:pmlb-classification}
\centering
\begin{tabular}{||c c c||} 
 \hline
 metric $\backslash$ method & SVE & TOM \\ 
 \hline\hline
 mean test acc & 81.5 & 81.7 \\ [1ex] 
 mean test acc & 81.7 & 81.9 \\ [1ex] 
 \hline
\end{tabular}
\end{table}

In order to ensure that these results are not driven by the presence of UCI-121 datasets, we train both methods again on the $82$ tasks not present in UCI-121 (experiment repeated twice). The results are shown in Table \ref{tab:pmlb-classification-non-uci-121}.

\begin{table}[h]
\caption{Test set accuracy on the non-UCI-121 part of the PMLB Classification dataset (experiment repeated twice).}
\label{tab:pmlb-classification-non-uci-121}
\centering
\begin{tabular}{||c c c||} 
 \hline
 metric $\backslash$ method & SVE & TOM \\ 
 \hline\hline
 mean test acc & 79.9 & 80.4 \\ [1ex] 
 mean test acc & 79.6 & 78.6 \\ [1ex] 
 \hline
\end{tabular}
\end{table}

We see that SVE retains the classification power on par with TOM on both datasets using exactly the same setup as for UCI-121, i.e. no hyperparameter optimization is performed for the PMBL dataset. The test accuracy of SVE is slightly lower than that of the baseline, but our main goal is to show that the system with interpretable components performs on par with the baseline, which is supported by these results.

\section{Relation to VQ-VAE}
\label{app:vq-vae}

It is worth commenting on how SVE is related to VQ-VAE \cite{vandenOord2017}, a generative model with shared components.

SVE relies on the attention mechanism to combine the \textit{shared variable embeddings} based on a query \textit{raw variable embedding} to produce a \textit{processed variable embedding} for the input variables. The use of attention is ubiquitous in various areas of machine learning, however, we are not aware of the extensive use of attention for \textit{multi-task learning} (MTL) for tabular data. Additionally, SVE is not a straightforward application of attention to vector representations. The difference lies in the fact that for each input variable we obtain a tuple $(x_{i}, \mathbf{z}_{i})$, where $x_{i}$ is simply the value of the variable from the dataset and $\mathbf{z}_{i}$ is the actual raw variable embedding. It is to those raw embeddings that the attention mechanism is applied. The value $x_{i}$ remains untouched in the procedure, and the final tuple fed into the encoders/decoder is $(x_{i}, \mathbf{f}_{i})$, where $\mathbf{f}_{i}$ is the processed variable embedding. This means that our attention mechanism is tasked with obtaining the final \textit{name} of the variable from a set of learnable concepts - the shared embedding matrix. The \textit{value} of the variable is left as is.

There are a number of major differences between SVE and VQ-VAE and its extensions:
\begin{itemize}
    \item VQ-VAE is a generative model in the sense that it produces data which is intended to resemble data from the dataset. The SVE method is predictive rather generative in that it predicts the values of individual target variables.
    \item VQ-VAE is trained to reconstruct the input and it uses, among others, a reconstruction loss in its training objective. The SVE method is not trained to reconstruct the input at all but rather to predict values of variables not present in the input.
    \item After training, the VQ-VAE decoder can be used to generate samples from the input domain using randomness. The SVE method has no such mechanism and does not in any way attempt to produce data consistent with the input data domain.
    \item In VQ-VAE the decoder initially outputs a sequence of ordered representations which are then replaced by their closest counterparts in the codebook. In the SVE method, it is only the variable embedding, or the name of the variable, that is being replaced and the replacement is not with one vector from the shared embeddng matrix but rather with a linear combination of vectors from the shared embedding matrix. In our opinion, this difference alone is enough to differentiate the SVE method from VQ-VAE.
    \item Due to its use of the codebook, VQ-VAE is not fully differentiable and requires the use of a method akin to the straight-through estimator. The SVE method relies on a completely different access mechanism to the shared vectors and is end-to-end differentiable without any use of a straight-through estimator.
    \item The introduction of the quantization in VQ-VAE is done to restrict the latent space of the model by pinning down representations to elements from the codebook. The SVE method does not attempt to do this but rather is focused on expressing each input variable name in terms of a set of shared components.
    \item VQ-VAE is typically applied to vision tasks, while the SVE method is geared towards tabular data, or data which can be conveniently represented in tabular form. The representations obtained in VQ-VAE are \textit{localized} in the sense that they are tied to specific regions of the image from which they have been obtained. Such structure is not present in the the SVE formulation. More than that, the summing operator in Eq. \ref{eq:shared} specifically tells us that the order of the variables is not particularly important, other than in the VQ-VAE, where quanitzed features are tied to the patches of the image.
    \item Another difference comes from the quantization mechanism in VQ-VAE, where the input is mapped to an ordered sequence of representations, each of which is quantized from a shared set. While the codebook in VQ-VAE is unordered, the final representation to which the VQ-VAE encoder maps consists of a sequence of vectors from the codebook, where each vector corresponds to a representation of a part of the image. SVE does not directly map the input to quantized representations but rather allows the identifier or the \textit{name} of the variable to be recombined from a set of real-valued vectors which can be trained with SGD. Both of these differences highlight that careful consideration is required before applying shared embedding methods to the variable embedding setting.
\end{itemize}

\section{Motivation of increased number of parameters in SVE}
\label{app:parameters-increase}
Shared representations are a recurring theme in MTL - this is also true for many other areas of machine learning, and most of the research in deep learning. Common methods for approaching MTL reuse representations by sharing parts of the processing pipeline, e.g. parts of the neural network, while producing specialized representations for individual tasks, e.g. by introducing specialized encoder or decoder heads and this is by no means a new idea, even if we focus exclusively on deep modern architectures \cite{Kaiser2017}. What sets our method apart from most of the research in the area is that it does not rely on task-specific parts of the architecture. More concretely, we do not have any sort of task-specific encoder or decoder, but rather rely on encoding and decoding on the \textit{variable} level. This means that the same encoders/decoder setup can be used for a multitude of seemingly unrelated tasks. More than that, those tasks do not have to have matching input and output dimensions. In principle, the only task-specific part of our architecture is the set of raw variable embeddings linked with a given tasks. And it is precisely this part of the model that SVE attempts to abstract away by using a set of shared variable embeddings not related to any given variable or task. The introduced shared embeddings indeed do not limit the number of parameters relative to the baseline but they may very well limit the overall number of parameters relative to settings with separate task-specific encoders and decoders. Also for our choice of $C = D = 128$ from the paper, the number of parameters introduced by the shared embeddings is $~ 128^{2}$. If we consider the UCI-121 dataset, and focus exclusively on the variable embeddings, the introduction of shared variable embeddings results in the number of parameters increasing by $128 / 3490 \approx 3.7\%$ as we introduce $128$ shared embeddings with the same dimensionality as the initial $3\,490$ raw variable embeddings. If we consider not only the embeddings but also other model parameters, e.g. encoders/decoder weights, etc., then the overall relative increase in the number of weight is smaller still. For the sparse attention model, the introduction of the shared embeddings causes the number of model parameters to grow by around $1\%$. This increase is then motivated by the increased interpretability of the final model.

\section{Training time}
\label{app:training-time}
We have observed a decrease in training time for the sparse attention shared embedding method vs. the vanilla shared embedding. To perform a meaningful comparison between the sparse attention method and our vanilla method, it would be useful to remark on the hardware used. We train all our models on a single NVIDIA A100 with 128GB of RAM. All of the experiments listed in the paper or the extended results take under 24 hours to train. More specifically, the $1.05$-entmax method takes about $13.5$ hours to train. We have found that it is possible to further limit the training time by using dropout. For instance, for the sparse attention method on the UCI-121 dataset, the introduction of dropout with a rate of $0.1$ brings the training time down to around $9$ hours $15$ minutes, while achieving the test accuracy of $81.5\%$. The same method trained on the PMLB dataset has a training time of under $7$ hours $30$ minutes, with the test set accuracy at $81.7\%$. As far as theoretical complexity is concerned, attention itself has the complexity of $O(n^{2} \cdot d)$, where $n$ is the length of the sequence and $d$ is the dimensionality of the representation \cite{Vaswani2017}. This fact itself does not turn out to be a problem in practice in many cases, e.g. in NLP, and it is not a problem in our specific case.

\section{Variable embeddings}
\label{app:variable-embeddings}
The reason behind the introduction of variable embeddings follows a line of investigation starting in other domains, most notably NLP. First, \textit{word embeddings} were introduced \cite{Bengio2000} as a way to provide distributed, learnable representations for words in a vocabulary, which could then be used by a language model. This has led to work on actually embedding the representation vector in the contexts in which it occurs in the data \cite{Mikolov2013} and this is where the term \textit{embedding} comes from. Once it was shown that word embeddings performed well on NLP tasks, they started to serve as an inspiration for other domains. In the MTL domain, they inspired the introduction of \textit{task embeddings}, which provided descriptions or \textit{names} for tasks \cite{Yang2014, Zintgraf2019}, allowing more general models to operate on different tasks by receiving a task embedding vector. Embeddings were then introduced on the level of individual variables rather than individual tasks \cite{Meyerson2021}. This allowed the model to be agnostic to the number of input and output variables, which translates into one model being able to handle problems with significantly different input/output dimensionality. Empirically, it turns out that for tabular data such a model has significantly higher predictive power than using an ensemble of individually trained models or a model with a general core component but task-specialized encoders/decoders.

\section{Hinge loss}
\label{app:hinge-loss}
The choice of hinge loss over other possible loss functions, e.g. cross-entropy loss, was dictated by the same rationale as our choice of hyperparameters - the decision to provide a direct comparison with the TOM baseline. Since TOM uses a squared hinge loss, this was the loss that we adopted to tease out the effect of introducing the shared embeddings. As far as the squared hinge loss itself is concerned, it is more suitable than cross-entropy in this particular setting as it does not require passing the output through a softmax activation, which allows the individual components of the output to remain separate. An additional reason is that for very good predictions the loss hits zero, other than is the case for cross-entropy. This prevents the model from overfitting on already well-predicted samples.

\section{Code and data}
\label{app:code}
The code required to reproduce the experiments described in this paper is uploaded as supplementary material. The dataset used to train the models is a version of the UCI-121 dataset \cite{Delgado2014, Kelly2023}, with custom preprocessing. It is available publicly at \url{https://drive.google.com/file/d/1Wtq0hFxmO2INs0TxYmBP_aayEjjDZlJr/view?usp=drive_link}.

\end{document}